\documentclass[dvipsnames,format=sigconf,anonymous=false,review=false,pbalance=true,nonacm]{acmart}

\usepackage[utf8]{inputenc}
\usepackage{amsmath,mathtools,bbm} 
\usepackage[inline,shortlabels]{enumitem}
\usepackage[detect-weight=true]{siunitx}
\PassOptionsToPackage{hyphens}{url}\usepackage{hyperref}
\usepackage{cleveref}
\usepackage{algorithm}
\usepackage[]{algpseudocode}
\crefname{algocf}{alg.}{algs.}
\Crefname{algocf}{Algorithm}{Algorithms}
\usepackage{color,colortbl,etoolbox}
\robustify\bfseries
\usepackage{multirow}
\usepackage{subcaption}
\usepackage{csquotes} 
\usepackage[normalem]{ulem}
\usepackage{afterpage}
\usepackage{booktabs}
\renewcommand\vec{\boldsymbol}

\DeclareMathOperator*{\argmax}{arg\,max}
\renewcommand\mod{\ \mathrm{mod}\ }
\usepackage{stackengine}

\usepackage{glossaries}
\usepackage{adjustbox}

\usepackage{fontawesome5}

\usepackage{tikz}
\usetikzlibrary{external,fit,positioning,backgrounds}
\usepackage[linewidth=0.75pt,ticklabelscale=0.75,labelscale=1,significancemarksscale=0.75]{plotmacros}

\usepackage[basecolor=gray,size=1mm,nOfSensorTypes=5]{vsrs}

\glsdisablehyper

\usepackage{xspace}
\def\ie{\emph{i.e.}\xspace}
\def\Ie{\emph{I.e.}\xspace}
\def\eg{e.g.\xspace}

\def\vs{vs.\xspace}

\colorlet{col.individual}{gray}
\colorlet{col.parent}{col17}
\colorlet{col.best1}{col11}
\colorlet{col.best8}{col15}
\colorlet{col.similar1}{col12}
\colorlet{col.similar8}{col13}
\colorlet{col.random1}{col14}
\colorlet{col.random8}{col18}
\colorlet{col.random}{black}
\readlist\colorlist{col.random,col.individual,col.parent,col.best1,col.best8,col.similar1,col.similar8,col.random1,col.random8}

\newacronym{vsr}{VSR}{virtual soft robot}
\newacronym{ec}{EC}{evolutionary computation}
\newacronym{er}{ER}{evolutionary robotics}
\newacronym{ga}{GA}{genetic algorithm}
\newacronym{ea}{EA}{evolutionary algorithm}
\newacronym{bo}{BO}{Bayesian optimization}
\newacronym{sl}{SL}{social learning}
\newacronym{il}{IL}{individual learning}
\newacronym{ann}{ANN}{artificial neural network}
\newacronym{mlp}{MLP}{multilayer perceptron}
\newacronym{qd}{QD}{quality-diversity}

\newcommand{\preprintnotice}{%
\begingroup
\renewcommand\thefootnote{}\footnote{%
Accepted at the Genetic and Evolutionary Computation Conference (GECCO 2026).\\
This is the author's version of the work. The final version will be available via the ACM Digital Library.%
}
\addtocounter{footnote}{-1}
\endgroup
}

\begin{document}

\title[\Acrlong{sl} for Evolved \acrshortpl{vsr}]{Social Learning Strategies for Evolved Virtual Soft Robots}

\author{K.~Ege de Bruin}
\email{eged@ifi.uio.no}
\orcid{0000-0003-0111-7078}
\affiliation{
    \institution{Department of Informatics, University of Oslo}
    \city{Oslo}
    \country{Norway}
}

\author{Kyrre Glette}
\email{kyrrehg@ifi.uio.no}
\orcid{0000-0003-3550-3225}
\affiliation{
    \institution{Department of Informatics, University of Oslo}
    \city{Oslo}
    \country{Norway}
}
\additionalaffiliation{
    \institution{RITMO, University of Oslo}
    \city{Oslo}
    \country{Norway}
}

\author{Kai Olav Ellefsen}
\email{kaiolae@ifi.uio.no}
\orcid{0000-0003-2466-2319}
\affiliation{
    \institution{Department of Informatics, University of Oslo}
    \city{Oslo}
    \country{Norway}
}

\author{Giorgia Nadizar}
\email{giorgia.nadizar@irit.fr}
\orcid{0000-0002-3535-9748}
\affiliation{
    \institution{University Toulouse Capitole, IRIT - CNRS UMR5505}
    \city{Toulouse} 
    \country{France} 
}

\author{Eric Medvet}
\email{emedvet@units.it}
\orcid{0000-0001-5652-2113}
\affiliation{
    \institution{University of Trieste}
    \city{Trieste}
    \country{Italy}
}

\begin{abstract}
    Optimizing the body and brain of a robot is a coupled challenge: the morphology determines what control strategies are effective, while the control parameters influence how well the morphology performs. 
    This joint optimization can be done through nested loops of evolutionary and learning processes, where the control parameters of each robot are learned independently. 
    However, the control parameters learned by one robot may contain valuable information for others. 
    Thus, we introduce a social learning approach in which robots can exploit optimized parameters from their peers to accelerate their own brain optimization.
    Within this framework, we systematically investigate how the selection of teachers, deciding \emph{which} and \emph{how many} robots to learn from, affects performance, experimenting with virtual soft robots in four tasks and environments. 
    In particular, we study the effect of inheriting experience from morphologically similar robots due to the tightly coupled body and brain in robot optimization.
    Our results confirm the effectiveness of building on others' experience, as social learning clearly outperforms learning from scratch under equivalent computational budgets. 
    In addition, while the optimal teacher selection strategy remains open, our findings suggest that incorporating knowledge from multiple teachers can yield more consistent and robust improvements.
\end{abstract}

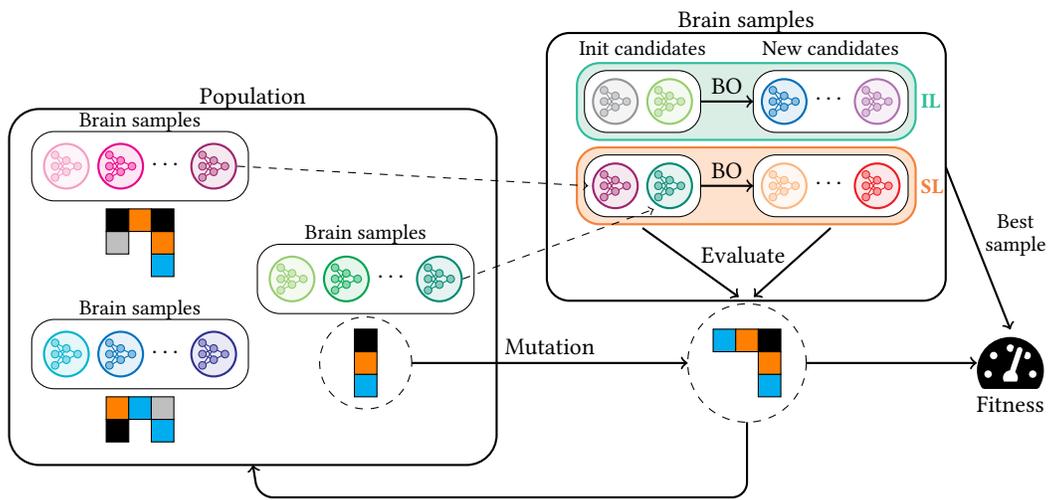
\begin{teaserfigure}
    \centering
    \pgfdeclarelayer{middle}   
\pgfsetlayers{background,middle,main}
\begin{tikzpicture}[
    node distance=.1cm, every node/.style={inner sep=0pt, outer sep=0pt},
    neuron/.style={circle, draw, fill, fill opacity=.4, minimum size=2mm, inner sep=0pt},
    edge/.style={draw},
    ann color/.style={
        /tikz/.cd,
        neuron/.append style={draw=#1},
        edge/.append style={draw=#1}
    },
    ann/.pic={
        \foreach \i/\y in {1/0.3, 2/0, 3/-0.3} {
          \node[neuron] (I\i) at (0,\y) {};
        }
        \foreach \i/\y in {1/0.15, 2/-0.15} {
          \node[neuron] (H\i) at (.3,\y) {};
        }
        \node[neuron] (O1) at (.6,0) {};
        \foreach \i in {1,2,3} {
          \foreach \j in {1,2} {
            \draw[edge] (I\i) -- (H\j);
          }
        }
        \foreach \j in {1,2} {
          \draw[edge] (H\j) -- (O1);
        }
        \node[draw, fill, fill opacity=0.1, circle, thick, inner sep=.1mm, fit=(I1)(I2)(I3)(H1)(H2)(O1)] {};
    }
]    
    \node (A1) {\tikz \pic[Turquoise, scale=0.5, transform shape]{ann};};
    \node[right=of A1] (A2) {\tikz \pic[RoyalBlue, scale=0.5, transform shape]{ann};};
    \node[right=of A2] (dots) {$\cdots$};
    \node[right=of dots] (A3) {\tikz \pic[Blue, scale=0.5, transform shape]{ann};};
    \node[draw,rounded corners=3mm,inner sep=1.5mm,fit=(A1)(A2)(dots)(A3),label=above:{\small{Brain samples}}] (wrapper) {};
    \node[below=of wrapper] (vsr1) {\vsrevogym[3mm]{3}{2}{342-104}};
    \node[yshift=2.5cm]  (B1) {\tikz \pic[Lavender, scale=0.5, transform shape]{ann};};
    \node[right=of B1] (B2) {\tikz \pic[Magenta, scale=0.5, transform shape]{ann};};
    \node[right=of B2] (dotsB) {$\cdots$};
    \node[right=of dotsB] (B3) {\tikz \pic[RedViolet, scale=0.5, transform shape]{ann};};
    \node[draw, rounded corners=3mm, inner sep=1.5mm, fit=(B1)(B2)(dotsB)(B3), label=above:{\small{Brain samples}}] (wrapper2) {};
    \node[below=of wrapper2] (vsr2) {\vsrevogym[3mm]{3}{3}{131-203-004}}; 
    \node[xshift=3cm, yshift=1cm]  (C1) {\tikz \pic[YellowGreen, scale=0.5, transform shape]{ann};};
    \node[right=of C1] (C2) {\tikz \pic[Green, scale=0.5, transform shape]{ann};};
    \node[right=of C2] (dotsC) {$\cdots$};
    \node[right=of dotsC] (C3) {\tikz \pic[PineGreen, scale=0.5, transform shape]{ann};};
    \node[draw, rounded corners=3mm, inner sep=1.5mm, fit=(C1)(C2)(dotsC)(C3), label=above:{\small{Brain samples}}] (wrapper3) {};
    \node[below=2mm of wrapper3] (vsr3) {\vsrevogym[3mm]{1}{3}{1-3-4}};
    \node[draw, dashed, circle, fit=(vsr3), inner sep=1mm] (circ1) {};
    \node[draw,rounded corners=4mm,thick,inner sep=3mm,fit=(wrapper)(wrapper2)(wrapper3)(vsr1)(vsr2)(vsr3),label=above:{Population}] (population) {};
    \node[right=4cm of circ1] (vsr4) {\vsrevogym[3mm]{3}{3}{431-003-004}}; 
    \node[draw, dashed, circle, fit=(vsr4), inner sep=1mm] (circ2) {};
    \draw[->,thick] (circ1) -- (circ2) node[midway, above, yshift=.2mm, inner sep=1mm] {Mutation};
    \node (L1) at ([xshift=-1.75cm, yshift=1.9cm]vsr4.north) {\tikz \pic[RedViolet, scale=0.5, transform shape]{ann};};
    \node[right=of L1] (L2) {\tikz \pic[PineGreen, scale=0.5, transform shape]{ann};};
    \draw[->, dashed] (B3.east) -- ([xshift=-.5mm]L1.west);
    \draw[->, dashed] (C3.east) -- ([xshift=1mm]L2.south west);
    \node[right=of L2,xshift=0.8cm] (R1) {\tikz \pic[Apricot, scale=0.5, transform shape]{ann};};
    \node[right=of R1] (dots) {$\cdots$};
    \node[right=of dots] (R2) {\tikz \pic[Red, scale=0.5, transform shape]{ann};};
    \begin{pgfonlayer}{middle}
        \node[draw, fill=white,rounded corners=3mm, inner sep=1mm, fit=(R1)(R2)] (fin_wrapper) {};
        \node[draw, fill=white,rounded corners=3mm, inner sep=1mm, fit=(L1)(L2)] (init_wrapper) {};
    \end{pgfonlayer}
    \begin{pgfonlayer}{background}
        \node[draw=Orange, thick,fill=Orange!20, rounded corners=3mm, inner sep=1mm, fit=(init_wrapper)(fin_wrapper), label=right:{\small{\ \textcolor{Orange}{\textbf{SL}}}}] (sl_wrap) {};
    \end{pgfonlayer}
    \node[inner sep=0pt] (evaluate_wrapper) at ([yshift=-6mm]fin_wrapper.south) {}; 
    \draw[->, thick] (init_wrapper.east) -- (fin_wrapper.west) node[midway, above,yshift=1mm] {BO}; 
    \node[above=of L1,yshift=0.4cm] (Z1) {\tikz \pic[Gray, scale=0.5, transform shape]{ann};};
    \node[right=of Z1] (Z2) {\tikz \pic[YellowGreen, scale=0.5, transform shape]{ann};};
    \node[right=of Z2,xshift=0.8cm] (Z3) {\tikz \pic[NavyBlue, scale=0.5, transform shape]{ann};};
    \node[right=of Z3] (dotsup) {$\cdots$};
    \node[right=of dotsup] (Z4) {\tikz \pic[Orchid, scale=0.5, transform shape]{ann};};
    \begin{pgfonlayer}{middle}
        \node[draw, rounded corners=3mm,fill=white, inner sep=1mm, fit=(Z1)(Z2), label={[yshift=2mm]above:{\small{Init candidates}}}] (z1_wrap) {};
        \node[draw, rounded corners=3mm,fill=white, inner sep=1mm, fit=(Z3)(Z4), label={[yshift=2mm]above:{\small{New candidates}}}] (z2_wrap) {};
    \end{pgfonlayer}
    \begin{pgfonlayer}{background}
        \node[draw=SeaGreen, thick,fill=SeaGreen!20, rounded corners=3mm,inner sep=1mm,fit=(z1_wrap)(z2_wrap),label=right:{\small{\ \textcolor{SeaGreen}{\textbf{IL}}}}] (il_wrap) {};
    \end{pgfonlayer}
    \draw[->, thick] (z1_wrap.east) -- (z2_wrap.west) node[midway, above,yshift=1mm] {BO}; 
    \node[draw, rounded corners=3mm, thick,inner sep=5mm,fit=(init_wrapper)(z2_wrap)(evaluate_wrapper), label=above:{{Brain samples}}] (wrapper_mutated_brain) {};
    \draw[->, thick] ([yshift=-1.5mm]init_wrapper.south) -- ($(wrapper_mutated_brain.south)+(-1mm,0.5mm)$) node[midway, above,xshift=7mm] {Evaluate}; 
    \draw[->, thick] ([yshift=-1.5mm]fin_wrapper.south) -- ($(wrapper_mutated_brain.south)+(1mm,0.5mm)$);     
    \node[right=of vsr4, yshift=0cm, xshift=2.5cm] (perf_icon) {\scalebox{2.5}{\faIcon{tachometer-alt}}};
    \node[below=1mm of perf_icon] {Fitness};
    \draw[->,thick] (circ2.east) -- (perf_icon.west);
    \draw[->,thick] (wrapper_mutated_brain.east) -- ([yshift=1mm]perf_icon.north)node[midway, above,yshift=-1mm,xshift=5mm] {\shortstack{\small Best \\ \small sample}};
    \draw[->, thick,rounded corners=2mm] (circ2.south) -- ++(0,-1) -- ++(-6.5,0) -- (population.south); 
\end{tikzpicture}
    \caption{
        Overview of the \gls{sl} approach. \normalfont 
        When a new \gls{vsr} is generated through mutation, we need to optimize its brain. 
        We use \acrlong{bo}, which works by using initial samples (\ie, initial \emph{evaluated} brains) to predict new candidates (\ie, new brains). 
        The best sample is then used as the \gls{vsr} brain to determine its fitness. 
        With \gls{sl}, initial candidates are obtained by inheriting samples from other \glspl{vsr} in the population.
        This differs from a standard case, \gls{il}, where initial candidates are independent from other robots.
        Several strategies exist for selecting which samples to inherit; in this figure we show inheritance of one sample from the most similar robots in the previous generation (we call this ``Similar-many'', see \Cref{sec:approach-brain}).
    }
    \label{fig:intro-picture}
\end{teaserfigure}


\maketitle
\preprintnotice
\glsresetall

\section{Introduction}
In \gls{er}, \glspl{ea} are used to optimize robot or virtual creature design and control~\cite{Sims1994,Lipson2000,Faina2013,Nolfi2016,Cheney2016}.
The \gls{ea} needs to traverse a more complex search space due to the intertwined body and brain~\cite{mertan2025evolutionary}.
Consequently, there might be a mismatch between control and robot morphology, and a robot morphology might be discarded because it has not been performing to its full potential due to poor control~\cite{Cheney2018,Luo2022,mertan2024investigating}. 

To deal with this problem, often morphology and controller optimization are separated from each other.
For example, this can be done with an \emph{outer evolutionary loop}, to optimize morphologies, and an \emph{inner learning loop}, where all morphologies go through a controller optimization phase~\cite{Eiben2013}.
Adding a learning loop is an efficient way to find well-performing robots~\cite{Miras2020,Luo2022,Gupta2021,pigozzi2023morphology}.
This is because new robot morphologies have more chance to reach more of their potential.
This does, however, come at the cost of requiring more resources to optimize a robot controller~\cite{Moreno2022,deBruin2025}.

If every new robot morphology goes through a controller-learn\-ing phase, one could follow a Darwinian approach, where control parameters are either randomly initialized for every robot morphology~\cite{Gupta2021}, or offspring robots inherit \emph{initial} control parameters from parents~\cite{Miras2020}.
However, this loses learned information new robots could use.
If a robot has good control parameters \emph{after} learning, these learned parameters might be useful for other robots as well.
This is where a Lamarckian approach, where optimized control parameters are transferred from parent to offspring, could be used.
In previous work, it has been shown that Lamarckian inheritance can improve performance~\cite{Jelisavcic2019,Luo2023,Harada2024}.
However, it is not necessary to limit the transfer of learned information from parent to offspring.
One can extend such transfer to \gls{sl}~\cite{Laland2004,Montes2008,Rendell2010,Heinerman2015,Bredeche2022}, where information can be exchanged from any robot to any other robot.
This is similar to a cultural algorithm, where information is centrally gathered and distributed~\cite{Reynolds1994,Hart2022}. 

In this work, we look into \gls{sl} for evolving \glspl{vsr} and compare \gls{sl} strategies.
We use \gls{bo} for controller optimization, as it provides a flexible framework for both \gls{il} and \gls{sl}.
Moreover, \gls{bo} has been shown to have a fast learning capability and sample efficiency~\cite{Chatzilygeroudis2019,Lan2021,VanDiggelen2024,deBruin2025,deBruin2025-2}.
In \gls{bo}, controller parameters, along with their observed quality (their objective value), are treated as samples.
Previous samples are then combined with a prior to predict the most promising candidate controller parameters to evaluate.
\gls{bo} uses the previously observed qualities and general uncertainty to predict promising candidates.
We define \gls{sl} as the transfer of samples from one or more robots to another robot, allowing a robot to learn from the samples of others rather than learning solely from its own samples.
One aspect we consider for selecting robots to learn from is morphological similarity.
Morphology and control are tightly coupled in embodied agents~\cite{Sims1994,Pfeifer2006}, such that morphologically similar robots often require similar control strategies for good performance~\cite{mertan2025controller,medvet2022impact}.
Therefore, controllers that perform well for one robot are likely to transfer effectively to morphologically similar robots.

The goal of this paper is to explore the role of \gls{sl} in \gls{er} and to better understand the conditions under which it is beneficial.
We consider how \gls{sl} mechanisms, such as similarity and performance, and how the number of teachers influence overall performance and robot design.
This will guide our study and provide a framework for analyzing how different forms of \gls{sl} interact with the dynamics of \gls{er}.

\section{Related work}

\subsection{Controller inheritance}
In previous work, it has been shown that adding controller optimization, also known as learning, for each morphology in \gls{er} can improve performance~\cite{Miras2020,Luo2022,deBruin2025}.
Controller optimization can be made more efficient by allowing optimized controller parameters to be exchanged between robots.
A common form of controller exchange is to inherit optimized controller parameters from parent to offspring, often described as \emph{Lamarckian inheritance}.
Studies have demonstrated the advantages of Lamarckian evolution over Darwinian evolution.
\citet{Jelisavcic2019} showed that inheriting optimized controllers, rather than initial populations, improves performance.
Similarly, \citet{Luo2023} found that passing optimized parameters to offspring enhances adaptation.
Moreover, \citet{Harada2024} use a transfer learning approach from parents to offspring where \gls{ann} weights are shared, and this improves performance over when weights are not shared.

It is not required to exchange information only from parent to offspring.
For other strategies, we distinguish between a \emph{cultural} approach and a \emph{social} approach.
In a \emph{cultural} approach, information among generations is centrally gathered and exchanged~\cite{Hart2022}.
\citet{LeGoff2022} keep an archive of learned controllers, where new robots only inherit learned control parameters from older robots with the same number of sensors and actuators.
\citet{Mertan2024a} teach a single controller using a diverse set of morphologies, which can then be used for the control of new morphologies.
In a \emph{social} approach, control parameters are directly transferred from robot to robot.
\citet{Heinerman2015} apply \gls{sl} to a population of morphologically identical robots, and show that adding \gls{sl} leads to better controllers faster, and that a combination of \gls{il} and \gls{sl} is beneficial.
Our work differs in that we also evolve and change morphology.
Finally, in a non-robotics artificial life setting, \citet{Bartoli2020} show that a small set of good teachers is most beneficial, and that the teacher, rather than the learner, should choose which samples to transfer.

\subsection{\Acrfull{bo}}
For optimizing robot controllers, we apply \gls{bo}~\cite{Mockus1974}, which is known for its sample efficiency, and its ability to balance the exploration-exploitation trade-off.
\gls{bo} is computationally expensive for high-dimensional problems with many samples, and it has been explored for morphology optimization in the same scenario (\ie, \glspl{vsr}) we are working on~\cite{Bhatia2021}.
In that study, \gls{bo} was evaluated and compared to other optimization methods for morphology optimization, and a \gls{ga} showed better performance.
It is shown by \citet{VanDiggelen2024} that \gls{bo} can perform well on robotic control optimization, outperforming an evolution strategy.
\citet{Lan2021} show that \gls{bo} works well at the start of a controller optimization process, but loses its advantage with more samples due to the previously mentioned computational complexity.
A Bayesian process can be improved by exploiting priors, where the process of controller learning can be kick-started using other information~\cite{Chatzilygeroudis2019}.
An example of this is using morphological parameters within the Bayesian controller optimization process~\cite{Liao2019}.
This allows the \gls{bo} process to take into account both body and controller characteristics when deciding where to explore next, and over generations this can improve controller optimization using morphological properties.
In our work, we do not use morphological parameters to improve the \gls{bo} process, instead we use controller information from previous robots in a similar manner as previous work~\cite{deBruin2025-2}.

\section{Background}
\label{sec:background}

\subsection{\Acrfullpl{vsr}}
\label{sec:background-vsrs}
For this study, we use as agents a kind of simulated modular soft robots known as \glspl{vsr} (also named voxel-based soft robots~\cite{hiller2011automatic,legrand2023reconfigurable,mertan2024investigating}).
A \gls{vsr} is defined by its morphology and its controller, also known as \emph{body} and \emph{brain}.

We simulate \glspl{vsr} in 2-D space and discrete time, using Evolution Gym~\cite{Bhatia2021}.

\subsubsection{\Gls{vsr} body.}
\label{sec:background-vsrs-body}
The body of a \gls{vsr} is described by a polyomino---a geometric shape composed of connected unit squares represented in a 2-D grid---of at most $5 \times 5$ units (voxels) linked together.
A voxel can be non-actuated (\vsrevogym[2mm]{1}{1}{1} or \vsrevogym[2mm]{1}{1}{2}) or actuated (\vsrevogym[2mm]{1}{1}{3} or \vsrevogym[2mm]{1}{1}{4}).
Non-actuated voxels can be rigid (\vsrevogym[2mm]{1}{1}{1}) or soft (\vsrevogym[2mm]{1}{1}{2}).
During a simulation, rigid voxels never change their square shape; soft voxels change their shape based on external forces applied on them (\eg, gravity or other voxels pushing them).
Actuated voxels can be horizontally actuated (\vsrevogym[2mm]{1}{1}{3}) or vertically actuated (\vsrevogym[2mm]{1}{1}{4}); both are soft.
During the simulation, the size of actuated voxels depends on external forces and an internal force that attempts to shrink or expand the voxel along the horizontal or vertical axis.
In our work, the relative deformation of an active voxel along the actuated axis is limited to the range $[0.6, 1.6]$.
The brain of the \gls{vsr} is in charge of determining the target deformation.

From now on, we denote by $B$ the set of \gls{vsr} bodies.

\subsubsection{\Gls{vsr} brain.}
\label{sec:background-vsrs-brain}
The brain of a \gls{vsr} is modular, distributed over the body, and based on \glspl{ann}~\cite{medvet2020evolution,Mertan2023}.
Namely, we use the architecture proposed in~\cite{Mertan2023}.
We put a \gls{mlp} in each voxel.
It takes as input:
\begin{enumerate*}[(a)]
    \item the current values of the horizontal velocity, vertical velocity, and voxel area of all the voxels within a Moore neighborhood of $1$,
    \item the horizontal and vertical distance to the closest object to the voxel (not considering other voxels of the \gls{vsr}),
    \item and a periodic time signal $k \mod 25$, $k$ being the simulation time step.
\end{enumerate*}
The input of the \gls{mlp} is hence defined in $\mathbb{R}^{30}$, with $30=3 \cdot 9 + 2 + 1$.
When no voxels are present in a given position of the neighborhood, we set the corresponding input values to $0$.
The output of the \gls{mlp} is a single number which we use directly for actuating the voxel, if actuated, after having rescaled it to $[0.6, 1.6]$.

We use the same architecture (\ie, number of layers), activation functions, and biases and synaptic weights for all the \glspl{mlp}.
Specifically, we use a single inner layer with $10$ neurons with ReLu as activation function, and we use the sigmoid function on the output neuron.
The brain of the \gls{vsr} can hence be described by the vector $\vec{\theta} \in \mathbb{R}^{321}$ of the biases and synaptic weights of the \glspl{mlp}.

We remark that, being composed of identical modules, this kind of brain is inherently transferable to any body, as the architecture of the \glspl{mlp} is independent from the shape of the body.

\subsection{\Acrfull{bo}}
\label{sec:background-bo}
\Gls{bo} is an iterative numerical optimization technique based on a surrogate model of the problem to be solved.

Let $f: \mathbb{R}^p \to \mathbb{R}$ be a black-box function, \ie, a function which can be applied (\emph{observed}), but whose concrete structure is unknown, and let $\argmax_{\vec{x} \in \mathbb{R}^p} f\left(\vec{x}\right)$ be a maximization problem based on $f$.

In brief, \gls{bo} works as follows.
Internally, it contains:
\begin{enumerate*}[(a)]
    \item a \emph{surrogate function} $\hat{f}$ of $f$ which produces estimates with uncertainty: given an $\vec{x} \in \mathbb{R}^p$, $\hat{f}_\mu(\vec{x})$ is the estimated value of $f(\vec{x})$ and $\hat{f}_\sigma\left(\vec{x}\right)$ is the uncertainty of the estimate;
    \item a procedure to obtain the surrogate from a set $\left\{\left(\vec{x}_i, f\left(\vec{x}_i\right)\right)\right\}_i$ of samples, which are pairs of candidate points and the corresponding observed values;
    \item an \emph{acquisition function} $a$ to obtain candidate points to evaluate next, using the surrogate function by means of $\bar{x}=\arg\max_{x\in\mathcal{X}}a(x; \hat{f})$.
\end{enumerate*}
Initially, \gls{bo} starts with a set $L_0=\left\{\vec{x}_i\right\}_{i=1}^{i=n_0}$ of $n_0$ initial candidate points---usually chosen according to a given probability distribution.
From this, it builds a set of samples $L'=\left\{\left(\vec{x}_i, f\left(\vec{x}_i\right)\right)\right\}_{i=1}^{i=n_0}$ by observing the value $f\left(\vec{x}_i\right)$ for each candidate point in $L_0$.
Then, it iterates the following steps until the number $|L'|$ of samples is equal to some predefined $n_\text{final}$:
\begin{enumerate*}[(i)]
    \item it updates the surrogate function $\hat{f}$ based on $L'$,
    \item it obtains a new candidate point $\bar{\vec{x}}$ from $\hat{f}$ using the acquisition function, 
    \item it observes the value $f\left(\bar{\vec{x}}\right)$ of $f$ at the new candidate point $\bar{\vec{x}}$, and
    \item it adds the sample $\left(\bar{\vec{x}}, f\left(\bar{\vec{x}}\right)\right)$ to $L'$.
\end{enumerate*}
At the end, \gls{bo} returns the sample $\left(\vec{x}^\star, f\left(\vec{x}^\star\right)\right) \in L'$ with the largest observed value.

The acquisition function plays a key role, as it balances the exploration-exploitation trade-off by attempting to produce a new candidate point $\bar{\vec{x}}$ that is, at the same time,
\begin{enumerate*}[(a)]
    \item good, \ie, with high $\hat{f}_\mu\left(\bar{\vec{x}}\right)$, and
    \item uncertain, \ie, with high $\hat{f}_\sigma\left(\bar{\vec{x}}\right)$, hence useful for improving the quality of the surrogate.
\end{enumerate*}
On the other hand, the surrogate computation and update have to be fast with respect to the computation of $f$.

There are several options for the various components of \gls{bo}.
In this work, we use the Matern 5/2 kernel with a length scale of $10$ as the surrogate function $\hat{f}$ and the upper confidence bound as the acquisition function, with an exploration variable of $3$, which we optimize using the L-BFGS-B algorithm~\cite{Zhu1997}.
These settings have been used and worked well for optimizing the controller of robots for directed locomotion~\cite{Lan2021,VanDiggelen2024,deBruin2025,deBruin2025-2}, and worked well in our preliminary experiments.

\subsubsection{\Gls{bo} as a form of learning.}
In the context of an agent interacting with an environment and attempting to perform a given task, \gls{bo} can be seen as a form of learning.

Let $\vec{\theta} \in \mathbb{R}^p$ be a parametrization of the policy of the agent, \ie, of its brain.
We can see $f$ as a way to measure the degree $f\left(\vec{\theta}\right)$ to which an agent with a given brain $\vec{\theta}$ successfully completes the task.
Applying $f$ requires deploying the agent in a real or simulated scenario and let it produce a behavior, out of which its quality $f\left(\vec{\theta}\right) \in \mathbb{R}$ is extracted.
Computing $f$ is usually very costly.

More broadly, different $\vec{\theta}_1,\vec{\theta}_2,\dots$ represent candidate parameter sets (the candidate points in the \gls{bo} process) for the brain and obtaining the respective observations $f\left(\vec{\theta}_1\right),f\left(\vec{\theta}_2\right),\dots$ corresponds to the agent interacting and gaining experience about the environment, \ie, learning.
In other words, a $\vec{\theta}$ is ``what to try'' and $f\left(\vec{\theta}\right)$ is ``what is the outcome''.
Based on this intuition, we show in the next section how we can use \gls{bo} to model \emph{social} learning by exploiting the fact that agents can ask other agents ``what to try'', possibly building on their experience and learning from it.

\section{\Acrlong{sl} with \gls{bo} for \glspl{vsr}}
\label{sec:approach}
We co-optimize the body and brain of \glspl{vsr} by using evolution for optimizing the bodies and \gls{bo} for optimizing the brains.
\Cref{fig:intro-picture} sketches our approach.

We assume that $q: B \times \mathbb{R}^{321} \to \mathbb{R}$ is the function to be maximized and corresponds to the task, representing the quality of candidate \gls{vsr}.
That is, $q\left(b, \vec{\theta}\right)$ measures how well a \gls{vsr} with a body $b \in B$ and a brain $\vec{\theta} \in \mathbb{R}^{321}$ performs the task.

\subsection{Body evolution}
\label{sec:approach-body}
For body evolution, we resort to a standard \gls{ga}.
We start with a population of $n_\text{pop}=200$ bodies obtained with random initialization (explained below).
Then, we repeat the following steps for $n_\text{gen}=50$ times (generations):
\begin{enumerate*}[(i)]
    \item we evaluate each body in the population;
    \item we generate an offspring of $n_\text{pop}$ new bodies, each one obtained by selecting one parent from the current population with tournament selection with $n_\text{tour}=4$ tournament size and then applying a mutation operator (see below);
    \item we replace all the parents with the offspring to obtain the new population.
\end{enumerate*}

For evaluating a body $b$, we pair it with a brain, which we optimize through learning, and assign to the body the quality $q\left(b,\vec{\theta}^\star\right)$ it obtained with the learned brain $\vec{\theta}^\star$.

During the evolution, we internally represent bodies as $5 \times 5$ matrices of elements defined in the alphabet of five symbols $V=\{\varnothing, \vsrevogym[2mm]{1}{1}{1}, \vsrevogym[2mm]{1}{1}{2}, \vsrevogym[2mm]{1}{1}{3}, \vsrevogym[2mm]{1}{1}{4}\}$, $\varnothing$ representing no voxel at a given position.
\Ie, the genotype space in our \gls{ga} is $V^{5\times5}$.
When randomly generating initial bodies, we proceed as follows:
\begin{enumerate*}[(i)]
    \item we set all the elements of the matrix to empty (\ie, $\varnothing$),
    \item we choose a random position in the matrix and a target number in $[10, 20] \subset \mathbb{N}$ of voxels for the body,
    \item we put a randomly chosen (\ie, $\sim U\left(\{\vsrevogym[2mm]{1}{1}{1}, \vsrevogym[2mm]{1}{1}{2}, \vsrevogym[2mm]{1}{1}{3}, \vsrevogym[2mm]{1}{1}{4}\}\right)$ voxel in the random position and iteratively add voxels in new positions adjacent with non-empty ones until reaching the target number of voxels;
\end{enumerate*}
this way, we obtain a valid polyomino of at least $10$ and at most $20$ voxels.
When mutating a matrix $\vec{v} \in V^{5\times5}$, we change one, two, or three random elements of $\vec{v}$ to a different symbol of $V$.
If the result is not a polyomino (\eg, there are more than one polyominoes in the matrix), or if the polyomino has less than $5$ voxels or more than $25$ voxels, we revert the mutation and try another one.
Other representations exist for \gls{vsr} bodies, \eg, more indirect~\cite{cheney2014unshackling} or grammar-based~\cite{megane2024grammar} ones: we decide to opt for this simple representation to keep the focus on the learning part.
Previous work also proved it to be effective~\cite{Mertan2023}.

\subsection{Brain learning}
\label{sec:approach-brain}
For learning the brain, we resort to \gls{bo}, which we cast as \gls{sl} by providing the initial candidate controller parameter sets based on samples from other \glspl{vsr}.
This works the same as the \emph{inherit samples} method in previous work~\cite{deBruin2025-2}, only instead of transferring samples only from parent to offspring, we compare different \gls{sl} strategies.
Recall that in the \gls{bo} process, a sample consists of a pair of a candidate point and its observed value, which in the \gls{vsr} context corresponds to a candidate set of controller parameters and its quality $q$.  
We compose the starting set $L_0$ of candidates by taking $n_0=8$ samples from the \glspl{vsr} of the previous generation and using their candidate controller parameter sets (at the first generation these candidates are random).
Then we execute \gls{bo} as explained in \Cref{sec:background-bo}, until reaching a number of $n_\text{final}=50$ samples.
Note that within the execution of \gls{bo}, $f\left(\vec{\theta}\right)$ is the degree of achievement of one body $b$ with the brain parametrization $\vec{\theta}$, \ie, $f\left(\vec{\theta}\right)=q\left(b, \vec{\theta}\right)$.

For collecting the $n_0$ initial candidates for a \gls{vsr} with body $b$, we experiment with seven approaches:
\begin{description}
    \item[Parent] We consider the parent of this \gls{vsr} and we take its best $n_0$ samples, \ie, the controllers with the largest observed value.
    \item[Best-One] We consider the best \gls{vsr}, \ie, the one with the largest $q$, of the previous generation and we take its best $n_0$ samples.
    \item[Best-Many] We consider the $n_0$ best \glspl{vsr} of the previous generation and we take, for each one, its best sample.
    \item[Similar-One] We consider the most morphologically similar \gls{vsr} to this $b$ in the previous generation and we take its best $n_0$ samples.
    \item[Similar-Many] We consider the $n_0$ the most morphologically similar \glspl{vsr} of the previous generation and we take, for each one, its best sample.
    \item[Random-One] We consider one random \gls{vsr} of the previous generation and we take its best $n_0$ samples.
    \item[Random-Many] We consider $n_0$ random \glspl{vsr} of the previous generation and we take, for each one, its best sample.
\end{description}
To measure the (dis)similarity between two bodies, we use the Hamming distance between the corresponding $5 \times 5$ matrices computed after having ``re-aligned'' the centers of mass of the two bodies.

It can be noted that the six variants differ along two axes:
\begin{enumerate*}[(a)]
    \item from which to learn from, \ie, who are the teachers (Best \vs Similar \vs Random), and
    \item from how many teachers to learn from (One \vs Many, \ie, $n_0$).
\end{enumerate*}
We remark that the amount of information the learner inherits is the same in all the cases, \ie, $n_0$ controller candidates.

\subsection{Baseline: \acrfull{il} with inheritance}
\label{sec:approach-baseline}
For providing a baseline for our experiments, we consider a further body-brain co-optimization scheme where the learning is individual, not social.
Here, \gls{bo} is still used for learning the brain, but the initial candidate is stored in the genotype and evolved with the body by \gls{ga}.

In detail, we set the genotype space of \gls{ga} to $V^{5 \times 5} \times \mathbb{R}^{321}$.
When we need to evaluate the quality of a \gls{vsr} $\left(b, \vec{\theta}\right)$, we use \gls{bo} with the initial $L_0 = \{\vec{\theta}\}$ and then we proceed as above.
When initially generating genotypes or mutating them, we operate as above for the $b$ part of the pair $\left(b, \vec{\theta}\right)$.
For $\vec{\theta}$, we generate initial values by randomly sampling in $[-1,1]^{321}$ and we do mutation by applying a perturbation sampled from the multivariate normal distribution $N(0, \sigma_\text{mut} \vec{I})$, $\vec{I} \in \mathbb{R}^{321 \times 321}$ being the identity matrix and $\sigma_\text{mut}=0.1$, \ie, we use the Gaussian mutation.

Note that in this form of optimization the initial candidate is inherited from the parent \gls{vsr} as genetic material, not inherited directly from the parent's experience, as in the Parent variant described in the previous section.
From another point of view, this form of optimization is Darwinian, whereas the Parent \gls{sl} variant is Lamarckian: both take something from the parent, but the former looks at the parent at its birth (as it takes the genetic material, which does not change during the parent life), while the latter looks at the parent at the end of its life (as it takes the best experiences it collected through its interactions with the environment).

\section{Experimental analysis}
\label{sec:exp}
We performed several experiments, \ie, optimization runs, for comparing the seven forms of \gls{sl} and the \gls{il} baseline.
We compared the approaches in terms of their ability to produce successful \glspl{vsr} in the task they are optimized for (\Cref{sec:exp-effectiveness}).
We analyzed in detail the morphological features of the optimized \glspl{vsr} (\Cref{sec:exp-bodies}).
We also assessed the generalization ability, \ie, whether the \glspl{vsr} the approaches produce can easily adapt to other tasks (\Cref{sec:exp-body-potential-transferability}).

For reference, data~\cite{dataset2026} and code for our experiments are available online at
\url{https://tinyurl.com/3rhzkm9t}.

\subsection{Tasks}
\label{sec:exp-tasks}
\begin{figure}
    \centering
    \begin{subfigure}{2cm}
        \centering
        \includegraphics[width=1\linewidth]{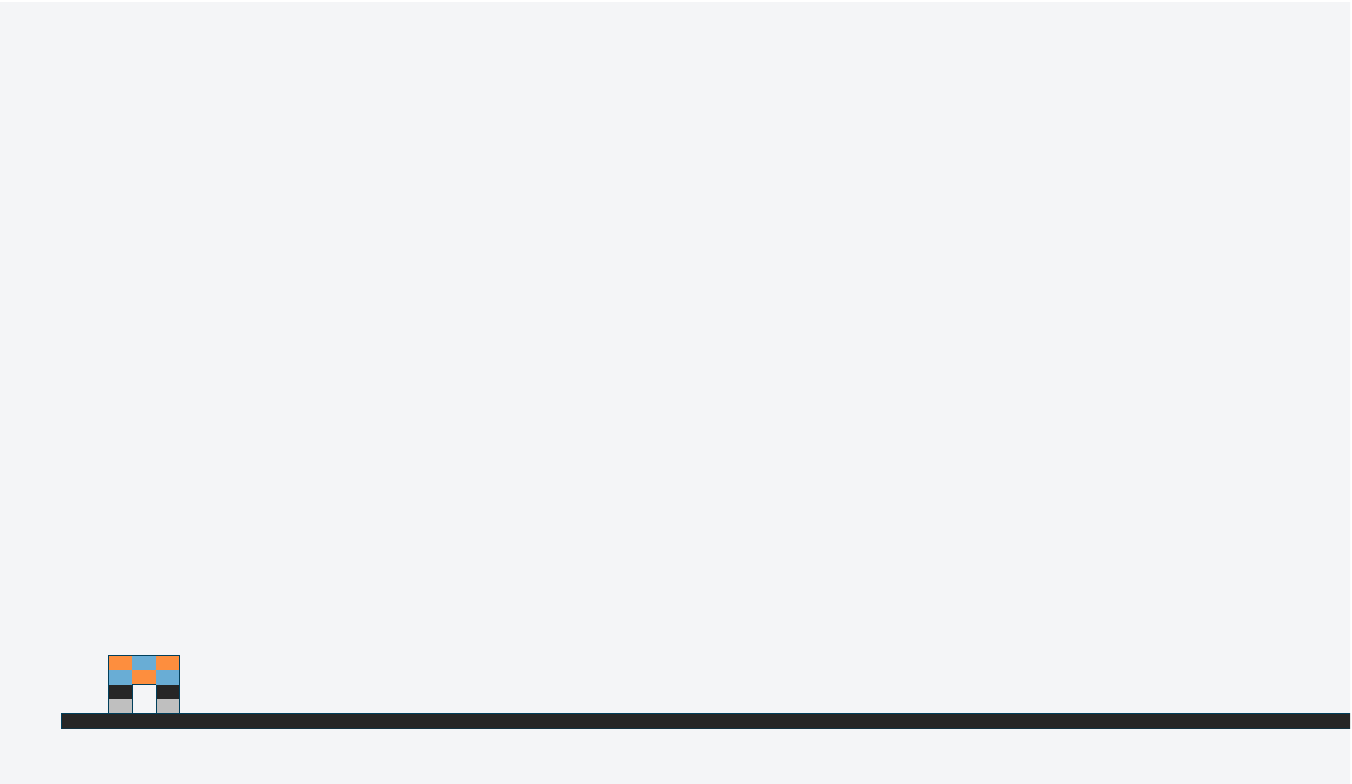}
        \caption{Simple}
    \end{subfigure}    
    \begin{subfigure}{2cm}
        \centering
        \includegraphics[width=1\linewidth]{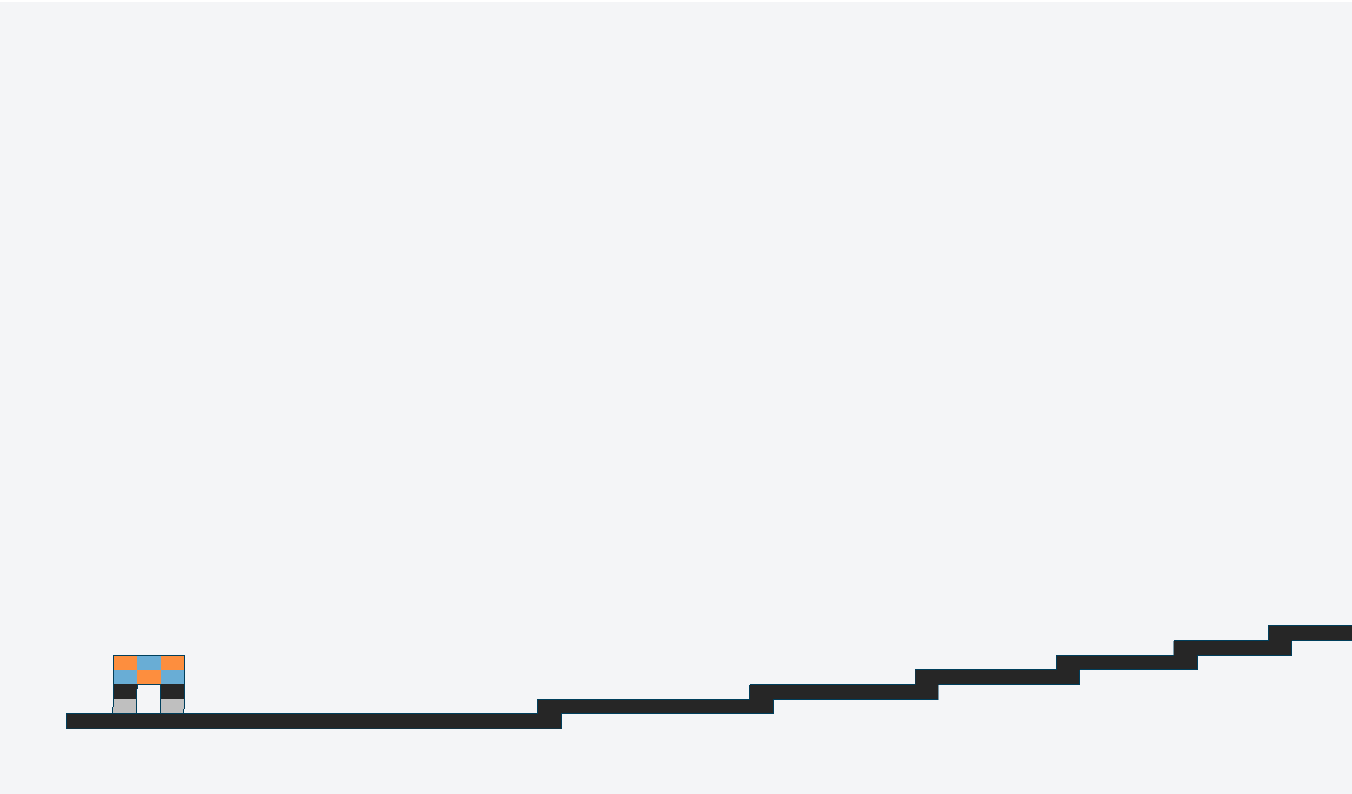}
        \caption{Steps}
    \end{subfigure}    
    \begin{subfigure}{2cm}
        \centering
        \includegraphics[width=1\linewidth]{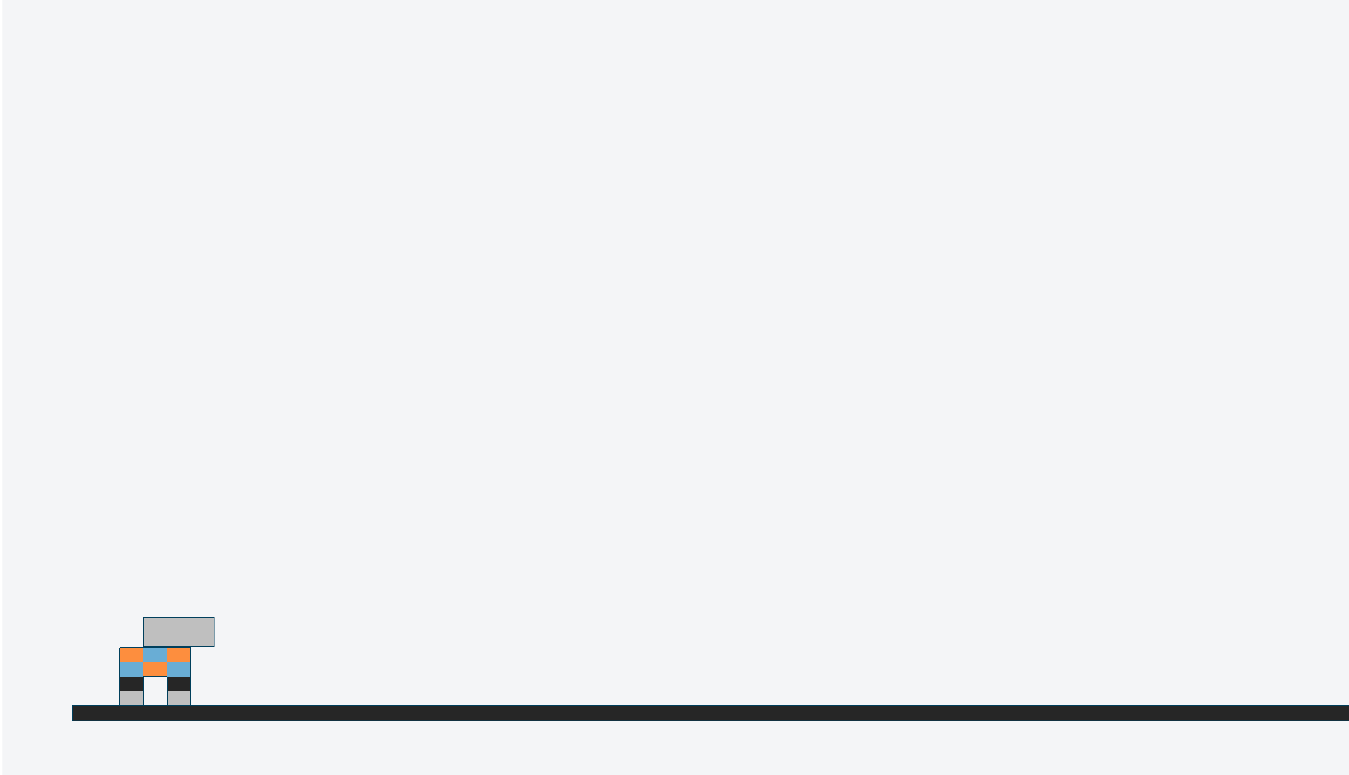}
        \caption{Carry}
    \end{subfigure}    
    \begin{subfigure}{2cm}
        \centering
        \includegraphics[width=1\linewidth]{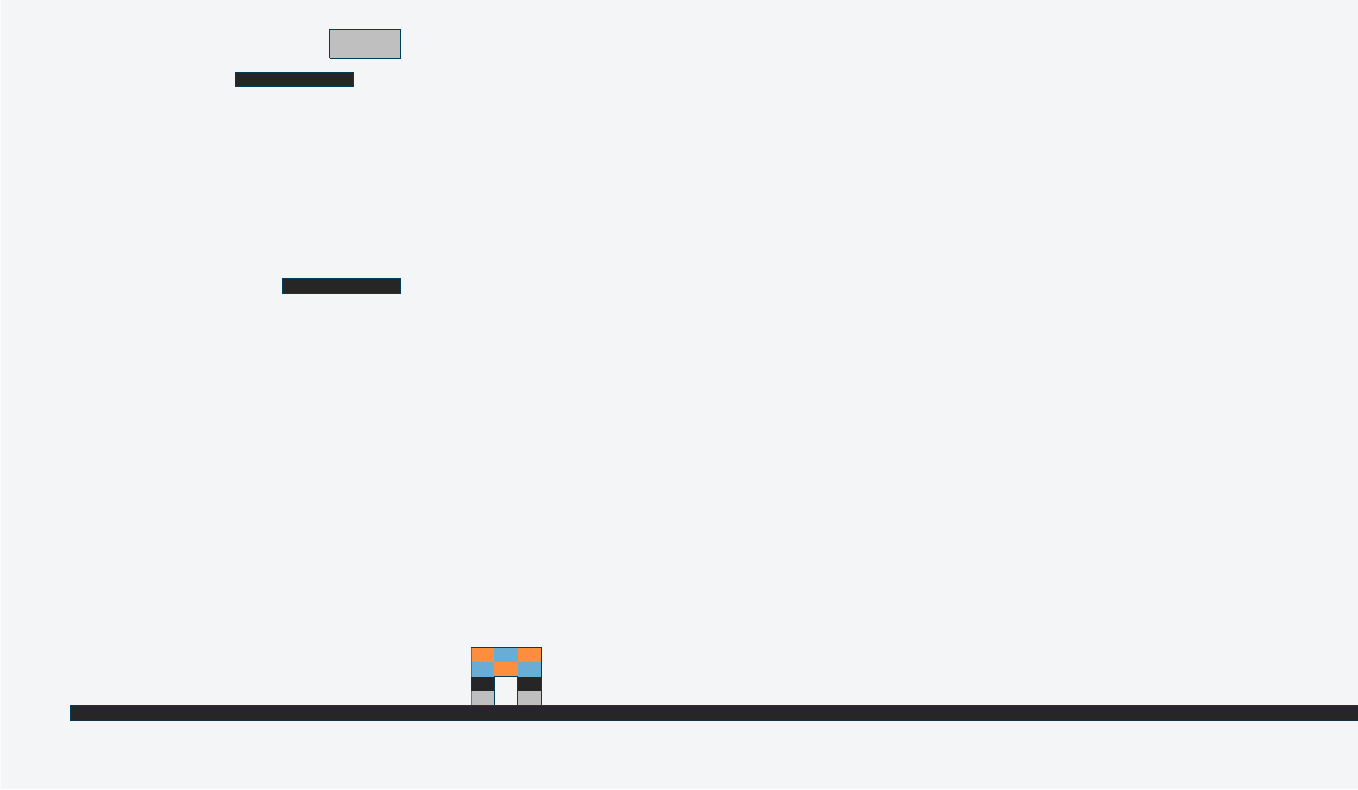}
        \caption{Catch}
    \end{subfigure}    
    \caption{
        A snapshot of four episodes where a \gls{vsr} performs the four tasks.
    }
    \label{fig:tasks}
\end{figure}

We consider four different tasks: Simple, Steps, Carry, and Catch.
\Cref{fig:tasks} shows the corresponding environments with a \gls{vsr} attempting to perform the task.

In \emph{Simple}, we put the \gls{vsr} on a flat surface and assign a quality $q$ to it which corresponds to the distance it run along the $x$-axis in the episode (lasting \num{500} time steps).
We compute the distance by considering the $x$-coordinate of the center of mass of the \gls{vsr} at the beginning ($k=0$) and at the end ($k=500$) of the episode.
This is a form of locomotion, a classic benchmark in \gls{er}.

\emph{Steps} is another, slightly harder, form of locomotion.
We place the \gls{vsr} on a stepped surface, instead of a flat one.
The quality is the same as the one of Simple.
For both Simple and Steps, consistent with previous works which used these tasks, we disabled the brain inputs providing the horizontal and vertical distance to the closest object to each voxel (see \Cref{sec:background-vsrs-brain}).

In the \emph{Carry} task, the \gls{vsr} is required to move while carrying an object.
We initially place the \gls{vsr} on a flat surface, as in Simple, but we additionally place an object (a rectangular box) on top of it.
The quality $q$ is the $x$-distance travelled by box, if it is still on top on the \gls{vsr} at the end of the episode, or the negative $x$-distance between the final position of the \gls{vsr} and the box, otherwise.
This way, we penalize behaviors where the \gls{vsr} drops its payload.

Finally, the \emph{Catch} task is a harder version of the Carry task in which we make the payload fall from a random position above the \gls{vsr} at the beginning of the episode.
Namely, the initial $x$-gap between the robot and box positions is not deterministic.
The \gls{vsr} is hence required to ``see'' the falling box, move to catch it, then move to carry it towards the right: the quality is the same as the one used in Carry.
Due to the varying initial position of the payload, this is a \emph{changing}, rather than static, environment: as such, it asks for a much greater generalization ability of the \glspl{vsr} than the previous three tasks.

\subsection{Effectiveness of the \gls{sl} approaches}
\label{sec:exp-effectiveness}
We applied the eight (seven \gls{sl} variants plus the \gls{il} baseline) approaches to the four tasks, \num{20} times for each combination.
To confirm the effectiveness of using \gls{bo} as a learning algorithm, for this analysis we also include another baseline, \emph{No-\gls{bo}}: in this variant, the $50$ controllers evaluated during the learning are chosen randomly, not using \gls{bo}---the final one used to assess the body is still the best one.
This results in a total of nine approaches to be compared.
We recall that we do $n_\text{gen}=50$ generations of a population of $n_\text{pop}=200$ \glspl{vsr} for the evolution of the body and evaluate $n_\text{final}=50$ controller parameter sets while optimizing the brain for each body.
It follows that we performed $9 \cdot 4 \cdot 20 \cdot 50 \cdot 200 \cdot 50 = \num{360000000}$ episodes.

When comparing approaches, we use the Mann-Whithney U test~\cite{Mann1947} for statistical significance analysis with the Benjamini-Hochberg correction~\cite{Benjamini1995} and a significance level of $\alpha=0.05$.

\begin{figure}
    \centering
    \begin{tikzpicture}
        \begin{groupplot}[
            boxplot,
            boxplot/draw direction=y,
            scale only axis,
            width=70mm,
            height=22.5mm,
            group style={
                group size=1 by 4,
                horizontal sep=7mm,
                vertical sep=5mm,
                xticklabels at=edge bottom,
                ylabels at=edge left
            },
            ygridded,
            noinnerticks,
            ylabel={Quality $q^\star$}
        ]
            \nextgroupplot[title={Simple}, ymax=185]
            \boxplotoutliers[bpcolor=col.random]{plot-data/5/5-performance-simple.txt}{Random};
            \boxplotoutliers[bpcolor=col.individual]{plot-data/5/5-performance-simple.txt}{Individual};
            \boxplotoutliers[bpcolor=col.parent]{plot-data/5/5-performance-simple.txt}{Parent};
            \boxplotoutliers[bpcolor=col.best1]{plot-data/5/5-performance-simple.txt}{Best-N=1};
            \boxplotoutliers[bpcolor=col.best8]{plot-data/5/5-performance-simple.txt}{Best-N=8};
            \boxplotoutliers[bpcolor=col.similar1]{plot-data/5/5-performance-simple.txt}{Similar-N=1};
            \boxplotoutliers[bpcolor=col.similar8]{plot-data/5/5-performance-simple.txt}{Similar-N=8};
            \boxplotoutliers[bpcolor=col.random1]{plot-data/5/5-performance-simple.txt}{Random-N=1};
            \boxplotoutliers[bpcolor=col.random8]{plot-data/5/5-performance-simple.txt}{Random-N=8};
            \boxplotcoloredstars{1}{170}{0,1,1,1,1,1,1,1,1}
            \boxplotcoloredstars{2}{162}{1,0,1,1,1,1,1,0,1}
            \boxplotcoloredstars{3}{170}{1,1,0,0,0,0,0,1,0}
            \boxplotcoloredstars{4}{162}{1,1,0,0,1,0,1,1,0}
            \boxplotcoloredstars{5}{170}{1,1,0,1,0,0,0,1,1}
            \boxplotcoloredstars{6}{162}{1,1,0,0,0,0,0,1,0}
            \boxplotcoloredstars{7}{170}{1,1,0,1,0,0,0,1,1}
            \boxplotcoloredstars{8}{162}{1,0,1,1,1,1,1,0,1}
            \boxplotcoloredstars{9}{170}{1,1,0,0,1,0,1,1,0}            
            
            \nextgroupplot[title={Steps}, ymax=85]
            \boxplotoutliers[bpcolor=col.random]{plot-data/5/5-performance-steps.txt}{Random};
            \boxplotoutliers[bpcolor=col.individual]{plot-data/5/5-performance-steps.txt}{Individual};
            \boxplotoutliers[bpcolor=col.parent]{plot-data/5/5-performance-steps.txt}{Parent};
            \boxplotoutliers[bpcolor=col.best1]{plot-data/5/5-performance-steps.txt}{Best-N=1};
            \boxplotoutliers[bpcolor=col.best8]{plot-data/5/5-performance-steps.txt}{Best-N=8};
            \boxplotoutliers[bpcolor=col.similar1]{plot-data/5/5-performance-steps.txt}{Similar-N=1};
            \boxplotoutliers[bpcolor=col.similar8]{plot-data/5/5-performance-steps.txt}{Similar-N=8};
            \boxplotoutliers[bpcolor=col.random1]{plot-data/5/5-performance-steps.txt}{Random-N=1};
            \boxplotoutliers[bpcolor=col.random8]{plot-data/5/5-performance-steps.txt}{Random-N=8};
            \boxplotcoloredstars{1}{80}{0,1,1,1,1,1,1,1,1}
            \boxplotcoloredstars{2}{77}{1,0,1,1,1,1,1,0,0}
            \boxplotcoloredstars{3}{80}{1,1,0,0,0,0,0,1,1}
            \boxplotcoloredstars{4}{77}{1,1,0,0,0,0,1,1,1}
            \boxplotcoloredstars{5}{80}{1,1,0,0,0,0,0,1,1}
            \boxplotcoloredstars{6}{77}{1,1,0,0,0,0,0,1,1}
            \boxplotcoloredstars{7}{80}{1,1,0,1,0,0,0,1,1}
            \boxplotcoloredstars{8}{77}{1,0,1,1,1,1,1,0,0}
            \boxplotcoloredstars{9}{80}{1,0,1,1,1,1,1,0,0}

            \nextgroupplot[title={Carry}, ymax=135]
            \boxplotoutliers[bpcolor=col.random]{plot-data/5/5-performance-carry.txt}{Random};
            \boxplotoutliers[bpcolor=col.individual]{plot-data/5/5-performance-carry.txt}{Individual};
            \boxplotoutliers[bpcolor=col.parent]{plot-data/5/5-performance-carry.txt}{Parent};
            \boxplotoutliers[bpcolor=col.best1]{plot-data/5/5-performance-carry.txt}{Best-N=1};
            \boxplotoutliers[bpcolor=col.best8]{plot-data/5/5-performance-carry.txt}{Best-N=8};
            \boxplotoutliers[bpcolor=col.similar1]{plot-data/5/5-performance-carry.txt}{Similar-N=1};
            \boxplotoutliers[bpcolor=col.similar8]{plot-data/5/5-performance-carry.txt}{Similar-N=8};
            \boxplotoutliers[bpcolor=col.random1]{plot-data/5/5-performance-carry.txt}{Random-N=1};
            \boxplotoutliers[bpcolor=col.random8]{plot-data/5/5-performance-carry.txt}{Random-N=8};
            \boxplotcoloredstars{1}{126}{0,1,1,1,1,1,1,1,1}
            \boxplotcoloredstars{2}{121}{1,0,1,1,1,1,1,0,1}
            \boxplotcoloredstars{3}{126}{1,1,0,0,0,0,0,1,0}
            \boxplotcoloredstars{4}{121}{1,1,0,0,0,0,0,1,0}
            \boxplotcoloredstars{5}{126}{1,1,0,0,0,0,0,1,0}
            \boxplotcoloredstars{6}{121}{1,1,0,0,0,0,0,1,0}
            \boxplotcoloredstars{7}{126}{1,1,0,0,0,0,0,1,0}
            \boxplotcoloredstars{8}{121}{1,0,1,1,1,1,1,0,0}
            \boxplotcoloredstars{9}{126}{1,1,0,0,0,0,0,0,0}

            \nextgroupplot[title={Catch}, ymax=110]
            \boxplotoutliers[bpcolor=col.random]{plot-data/5/5-performance-catch.txt}{Random};
            \boxplotoutliers[bpcolor=col.individual]{plot-data/5/5-performance-catch.txt}{Individual};
            \boxplotoutliers[bpcolor=col.parent]{plot-data/5/5-performance-catch.txt}{Parent};
            \boxplotoutliers[bpcolor=col.best1]{plot-data/5/5-performance-catch.txt}{Best-N=1};
            \boxplotoutliers[bpcolor=col.best8]{plot-data/5/5-performance-catch.txt}{Best-N=8};
            \boxplotoutliers[bpcolor=col.similar1]{plot-data/5/5-performance-catch.txt}{Similar-N=1};
            \boxplotoutliers[bpcolor=col.similar8]{plot-data/5/5-performance-catch.txt}{Similar-N=8};
            \boxplotoutliers[bpcolor=col.random1]{plot-data/5/5-performance-catch.txt}{Random-N=1};
            \boxplotoutliers[bpcolor=col.random8]{plot-data/5/5-performance-catch.txt}{Random-N=8};
            \boxplotcoloredstars{1}{100}{0,0,1,1,1,1,1,0,1}
            \boxplotcoloredstars{2}{96}{0,0,0,1,1,1,1,0,1}
            \boxplotcoloredstars{3}{100}{1,0,0,0,1,0,0,0,0}
            \boxplotcoloredstars{4}{96}{1,1,0,0,0,0,0,1,0}
            \boxplotcoloredstars{5}{100}{1,1,1,0,0,1,0,1,1}
            \boxplotcoloredstars{6}{96}{1,1,0,0,1,0,0,1,0}
            \boxplotcoloredstars{7}{100}{1,1,0,0,0,0,0,1,0}
            \boxplotcoloredstars{8}{96}{0,0,0,1,1,1,1,0,0}
            \boxplotcoloredstars{9}{100}{1,1,0,0,1,0,0,0,0}
            
        \end{groupplot}
    \end{tikzpicture}
    \begin{tabular}{rrr}
        No-\gls{bo} \addlegendimageintext{boxplot legend image={col.random}{}} &
        Best-One \addlegendimageintext{boxplot legend image={col.best1}{}} &
        Best-Many \addlegendimageintext{boxplot legend image={col.best8}{}} \\
        \gls{il} \addlegendimageintext{boxplot legend image={col.individual}{}} &
        Similar-One \addlegendimageintext{boxplot legend image={col.similar1}{}} &
        Similar-Many \addlegendimageintext{boxplot legend image={col.similar8}{}} \\
        Parent \addlegendimageintext{boxplot legend image={col.parent}{}} &
        Random-One \addlegendimageintext{boxplot legend image={col.random1}{}} &
        Random-Many \addlegendimageintext{boxplot legend image={col.random8}{}}
    \end{tabular}
    \caption{
         Distribution of the quality $q^\star$ of the best-performing \gls{vsr} in the four tasks for the eight approaches.
         Stars above the boxes show significant differences: for example, a red star on top of a gray box indicates that \gls{il} is significantly different from Best-One.
    }
    \label{fig:final-q-boxplots}
\end{figure}
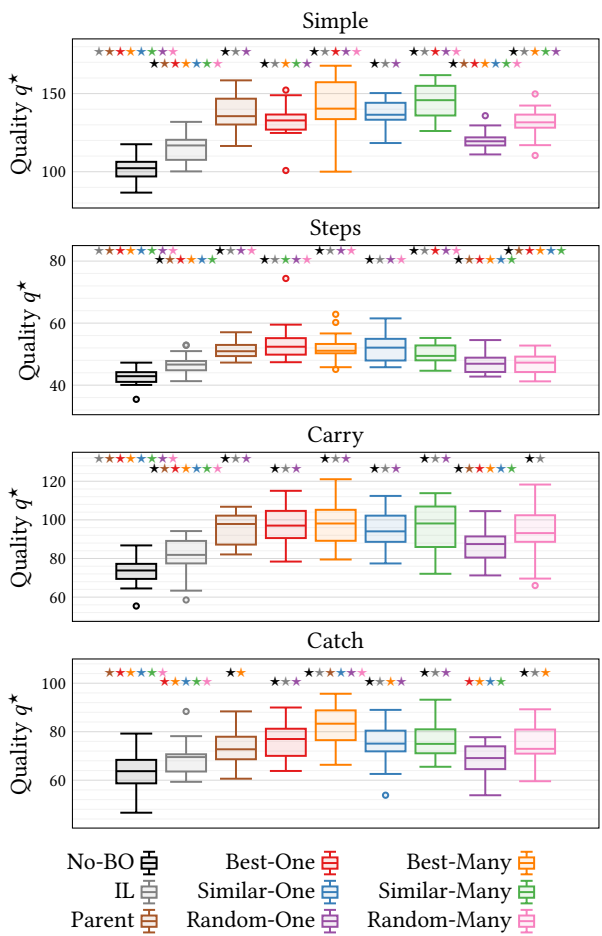

\Cref{fig:final-q-boxplots} presents the results for this first experiment.
It shows the distribution, across the $20$ repetitions, of the quality $q^\star$ of the best performing \gls{vsr} observed during the entire evolution with a given approach on a given task.
We recall that we use a generational model without overlapping, \ie, we drop the parents at each generation: for measuring $q^\star$ we hence retain the best \gls{vsr} across all the generations.

\Cref{fig:final-q-boxplots} firstly shows that all \gls{bo} methods outperform the No-\gls{bo} baseline, which confirms that \gls{bo} is efficient as a learning method.
Moreover, the figure also shows that, in general, \gls{sl} outperforms \gls{il}.
With the exception of Random-One, every \gls{sl} variant is significantly better than \gls{il} on at least three tasks.
Random-Many is better than \gls{il} on all but the Steps tasks.
All the remaining \gls{sl} variants are always better than \gls{il}.
This finding is sound: learning from a random teacher reveals to be not particularly better than having no teacher.
One observation is that in our dynamic \emph{Catch} task, we still see benefit of \gls{sl} over \gls{il}.
Related work outside robotics showed diminishing benefits of Lamarckian inheritance in dynamic environments~\cite{Paenke2007,Feldman1996,Sasaki1999,Ellefsen2013}.
Future work should look into the causes for this difference.

Another interesting observation we draw from \Cref{fig:final-q-boxplots} is that learning from more teachers is in general better than learning from just one teacher.
We recall that the ``amount of information'' we transfer from the teacher(s) to the learner is the same, as we keep $n_0=8$ in all \gls{sl} variants.
Despite the difference between the *-One and the *-Many variant is only occasionally statistically significant, this finding is sound and consistent with previous studies (\eg, \cite{Bartoli2020} observed a similar outcome, yet for a much simpler kind of agents).

\Cref{fig:final-q-boxplots} highlights no differences between Best-* and Similar-* variants.
We recall that choosing teachers based on morphological similarity is a peculiar opportunity enabled by our scenario, where we co-optimize the body and the brain.
The hypothesis that learning from a morphologically similar agent might bring more useful experience than learning from a good, yet possibly very different, teacher is not supported by our experiments.
We have not found clear motivations for this missed observation.
We hypothesize that the diversity of evolved bodies is not large enough to make Similar-* more convenient than Best-* (see next section).
We leave to future work the investigation of strategies where diversity is promoted (with a \gls{qd}~\cite{cully2017quality,nadizar2025enhancing} or a custom approach~\cite{pigozzi2023factors}) and teachers are chosen based on a combination of their quality and similarity.

\subsection{Analysis of the \gls{vsr} bodies}
\label{sec:exp-bodies}
We analyzed in deeper detail the bodies of the evolving \glspl{vsr}, as we wanted to verify whether different \gls{sl} variants lead to bodies evolving more or less diverse, or in different forms.

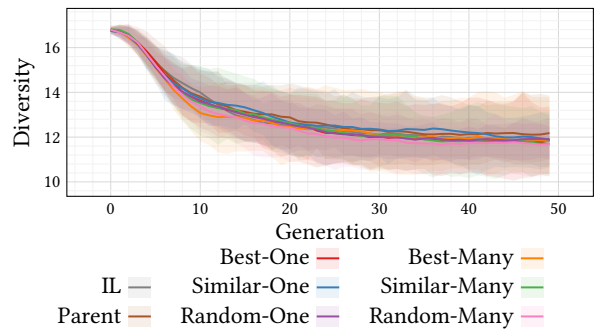
\begin{figure}
    \centering
    \begin{tikzpicture}[baseline=(current bounding box.center)]
        \begin{axis}[
            scale only axis,
            width=70mm,
            height=25mm,
            gridded,
            noinnerticks,
            xlabel={Generation},
            ylabel={Diversity}
        ]
            \lineminmax[lcolor=col.individual]{plot-data/7/7-diversity-none-0.txt}{}{x}{y}{ymin}{ymax}
            \lineminmax[lcolor=col.parent]{plot-data/7/7-diversity-parent-1.txt}{}{x}{y}{ymin}{ymax}
            \lineminmax[lcolor=col.best1]{plot-data/7/7-diversity-best-1.txt}{}{x}{y}{ymin}{ymax}
            \lineminmax[lcolor=col.best8]{plot-data/7/7-diversity-best-8.txt}{}{x}{y}{ymin}{ymax}
            \lineminmax[lcolor=col.similar1]{plot-data/7/7-diversity-similar-1.txt}{}{x}{y}{ymin}{ymax}
            \lineminmax[lcolor=col.similar8]{plot-data/7/7-diversity-similar-8.txt}{}{x}{y}{ymin}{ymax}
            \lineminmax[lcolor=col.random1]{plot-data/7/7-diversity-random-1.txt}{}{x}{y}{ymin}{ymax}
            \lineminmax[lcolor=col.random8]{plot-data/7/7-diversity-random-8.txt}{}{x}{y}{ymin}{ymax}            
        \end{axis}        
    \end{tikzpicture}    
    \begin{tabular}{rrr}
        &
        Best-One \addlegendimageintext{shaded legend image={col.best1}{}} &
        Best-Many \addlegendimageintext{shaded legend image={col.best8}{}} \\
        \gls{il} \addlegendimageintext{shaded legend image={col.individual}{}} &
        Similar-One \addlegendimageintext{shaded legend image={col.similar1}{}} &
        Similar-Many \addlegendimageintext{shaded legend image={col.similar8}{}} \\
        Parent \addlegendimageintext{shaded legend image={col.parent}{}} &
        Random-One \addlegendimageintext{shaded legend image={col.random1}{}} &
        Random-Many \addlegendimageintext{shaded legend image={col.random8}{}}
    \end{tabular}
    \caption{
        The evolution of the population diversity, averaged over runs and environments.
        The line corresponds to the mean value, the shaded area to the first-third quartile range.
    }
    \label{fig:diversity}
\end{figure}

We start by looking at the body diversity in the population during the evolution, which we show in \Cref{fig:diversity}.
For each evolutionary run and at each generation, we computed the average pairwise dissimilarity between bodies using the Hamming distance after re-alignment (see \Cref{sec:approach-brain})---the larger, the more diverse the population.

\Cref{fig:diversity} shows no difference among the eight approaches.
That is, regardless of the kind of learning (\gls{il} \vs \gls{sl}) or of the criterion for choosing teachers in \gls{sl}, \gls{vsr} bodies tend to be equally diverse, with a drop in diversity during the evolution likely caused by the evolutionary pressure.

We did a finer characterization of \gls{vsr} bodies by computing two simple descriptors (\ie, numerical features) for each one of them and visualizing the bodies in a plane defined by those two descriptors.
We measured the \emph{rate of active voxels} (\ie, the count of \vsrevogym[2mm]{1}{1}{3} and \vsrevogym[2mm]{1}{1}{4} divided by the count of \vsrevogym[2mm]{1}{1}{1}, \vsrevogym[2mm]{1}{1}{2}, \vsrevogym[2mm]{1}{1}{3}, and \vsrevogym[2mm]{1}{1}{4}) and the body \emph{compactness}, \ie, the ratio between the body area and the area of the convex hull.
We show the results of this analysis in \Cref{fig:body-features}, which contains one marker for the best \gls{vsr} optimized for every repetition, approach, and task.

\newcommand*\learnList{individual,parent,best1,best8,similar1,similar8,random1,random8}
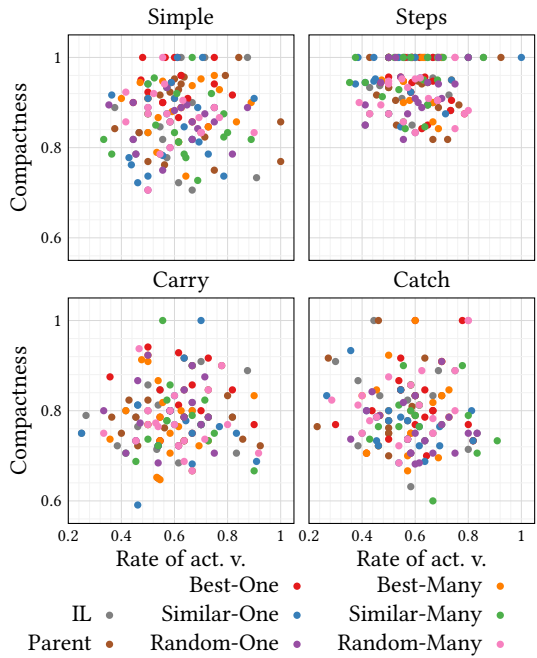
\begin{figure}
    \centering
    \begin{tikzpicture}
        \begin{groupplot}[
            scale only axis,
            width=30mm,
            height=30mm,
            group style={
                group size=2 by 2,
                horizontal sep=2mm,
                vertical sep=5mm,
                xticklabels at=edge bottom,
                xlabels at=edge bottom,
                yticklabels at=edge left,
                ylabels at=edge left
            },
            scale only axis,
            gridded,
            noinnerticks,
            xlabel={Rate of act.\ v.},
            ylabel={Compactness},
            xmin=0.2, xmax=1.05,
            ymin=0.55, ymax=1.05
        ]
            \nextgroupplot[title={Simple}]
            \expandafter\pgfplotsinvokeforeach\expandafter{\learnList}{\points[mcolor=col.#1,mtype=*]{plot-data/8/8-descriptors-simple.txt}{filter1=strategy is #1}{relative_activity}{compactness}}

            \nextgroupplot[title={Steps}]
            \expandafter\pgfplotsinvokeforeach\expandafter{\learnList}{\points[mcolor=col.#1,mtype=*]{plot-data/8/8-descriptors-steps.txt}{filter1=strategy is #1}{relative_activity}{compactness}}

            \nextgroupplot[title={Carry}]
            \expandafter\pgfplotsinvokeforeach\expandafter{\learnList}{\points[mcolor=col.#1,mtype=*]{plot-data/8/8-descriptors-carry.txt}{filter1=strategy is #1}{relative_activity}{compactness}}

            \nextgroupplot[title={Catch}]
            \expandafter\pgfplotsinvokeforeach\expandafter{\learnList}{\points[mcolor=col.#1,mtype=*]{plot-data/8/8-descriptors-catch.txt}{filter1=strategy is #1}{relative_activity}{compactness}}
        \end{groupplot}
    \end{tikzpicture}
    \begin{tabular}{rrr}
        &
        Best-One \addlegendimageintext{point legend image={col.best1}{*}{1pt}} &
        Best-Many \addlegendimageintext{point legend image={col.best8}{*}{1pt}} \\
        \gls{il} \addlegendimageintext{point legend image={col.individual}{*}{1pt}} &
        Similar-One \addlegendimageintext{point legend image={col.similar1}{*}{1pt}} &
        Similar-Many \addlegendimageintext{point legend image={col.similar8}{*}{1pt}} \\
        Parent \addlegendimageintext{point legend image={col.parent}{*}{1pt}} &
        Random-One \addlegendimageintext{point legend image={col.random1}{*}{1pt}} &
        Random-Many \addlegendimageintext{point legend image={col.random8}{*}{1pt}}
    \end{tabular}
    \caption{
        Body features for all the optimized \glspl{vsr}.
    }
    \label{fig:body-features}
\end{figure}

\begin{figure}
    \centering
    \adjincludegraphics[width=9mm,trim={{.25\width} {.25\width} {.25\width} {.25\width}},clip]{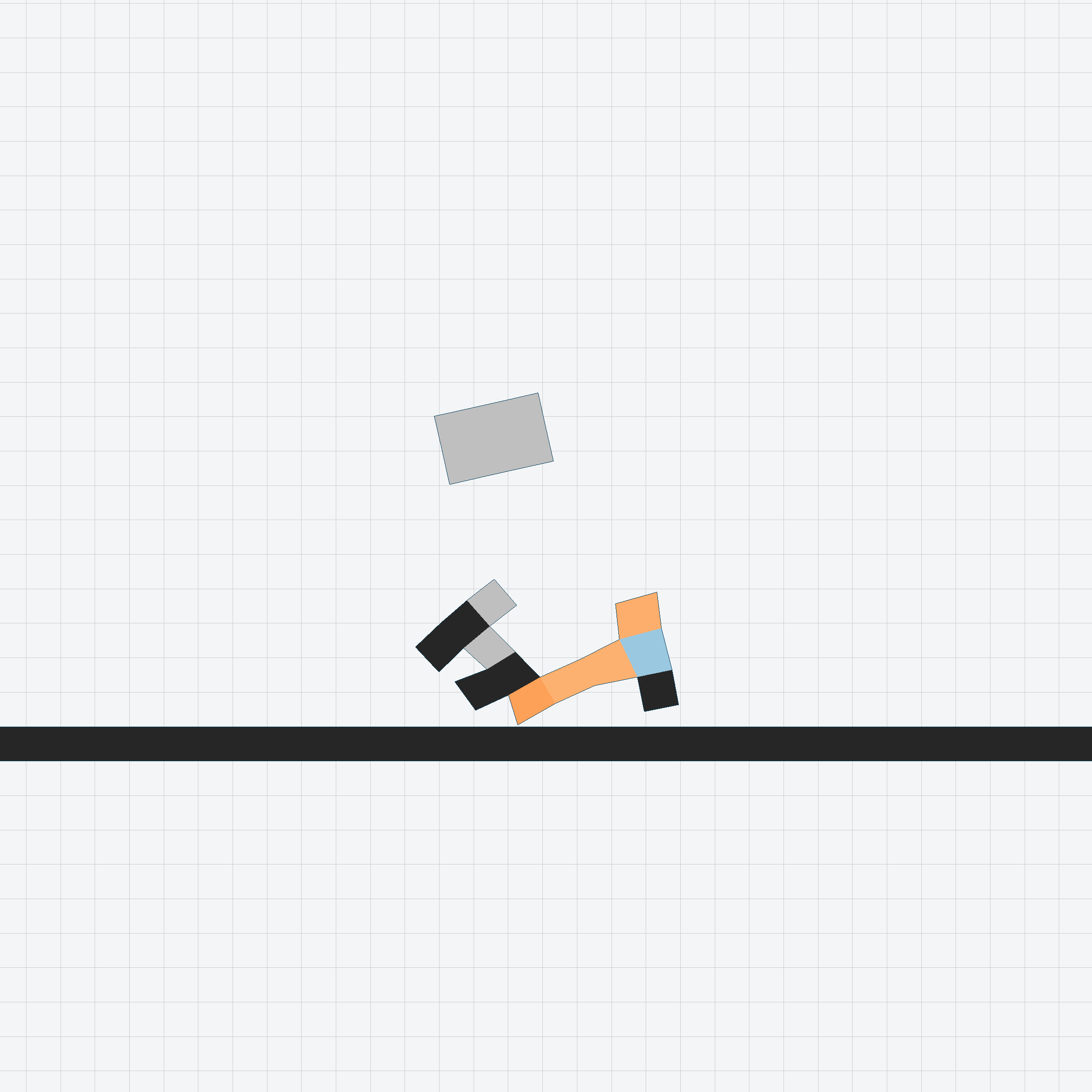}
    \adjincludegraphics[width=9mm,trim={{.25\width} {.25\width} {.25\width} {.25\width}},clip]{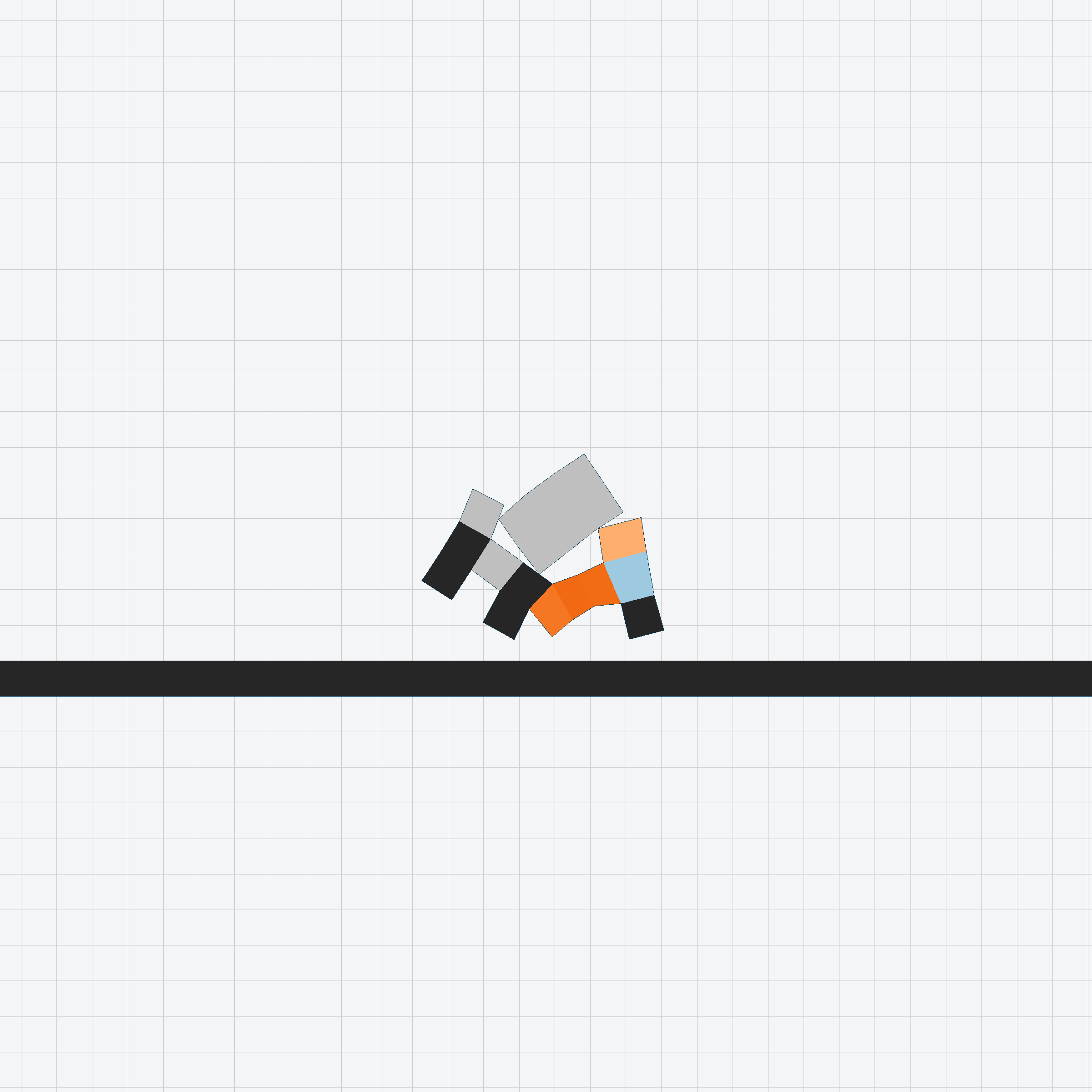}
    \adjincludegraphics[width=9mm,trim={{.25\width} {.25\width} {.25\width} {.25\width}},clip]{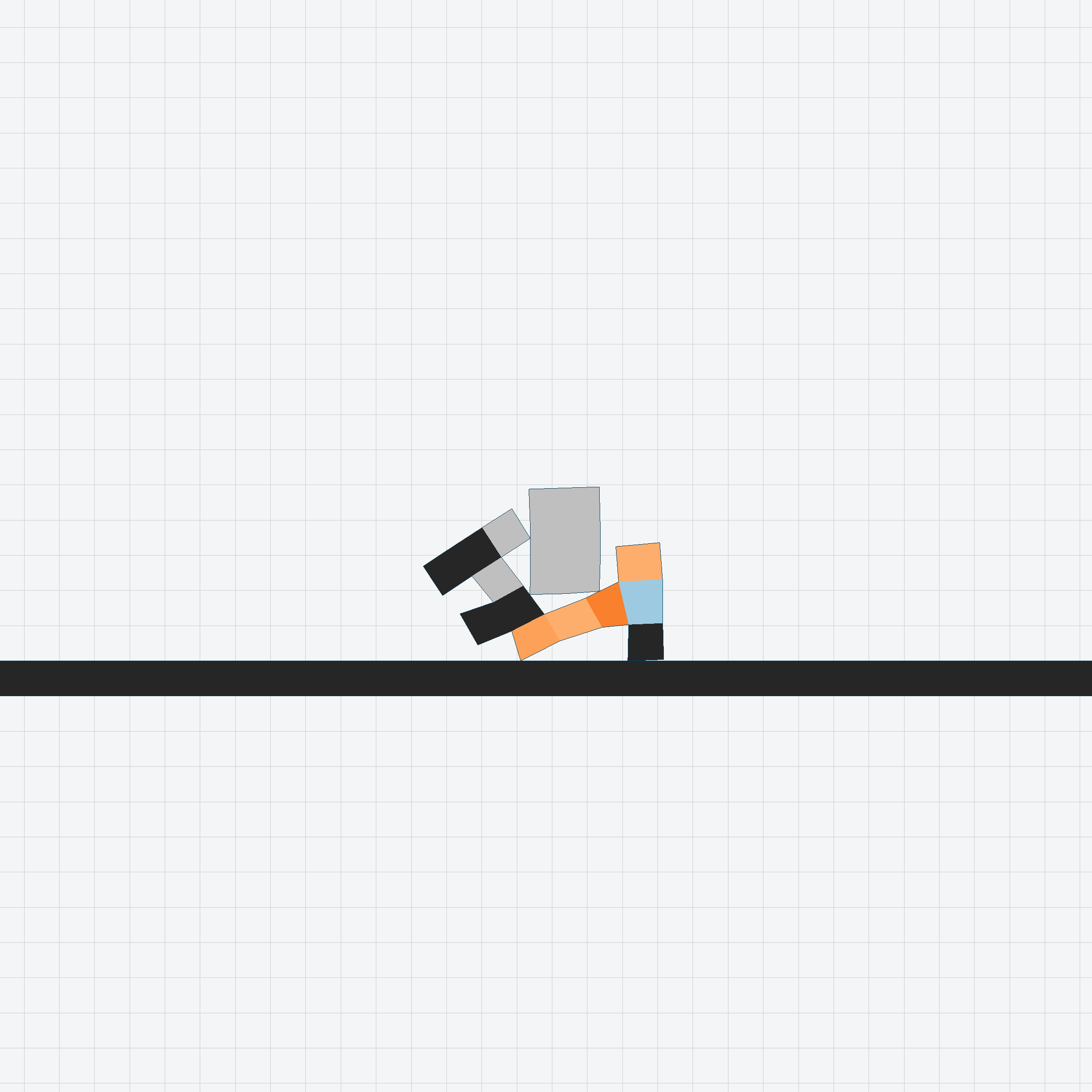}
    \adjincludegraphics[width=9mm,trim={{.25\width} {.25\width} {.25\width} {.25\width}},clip]{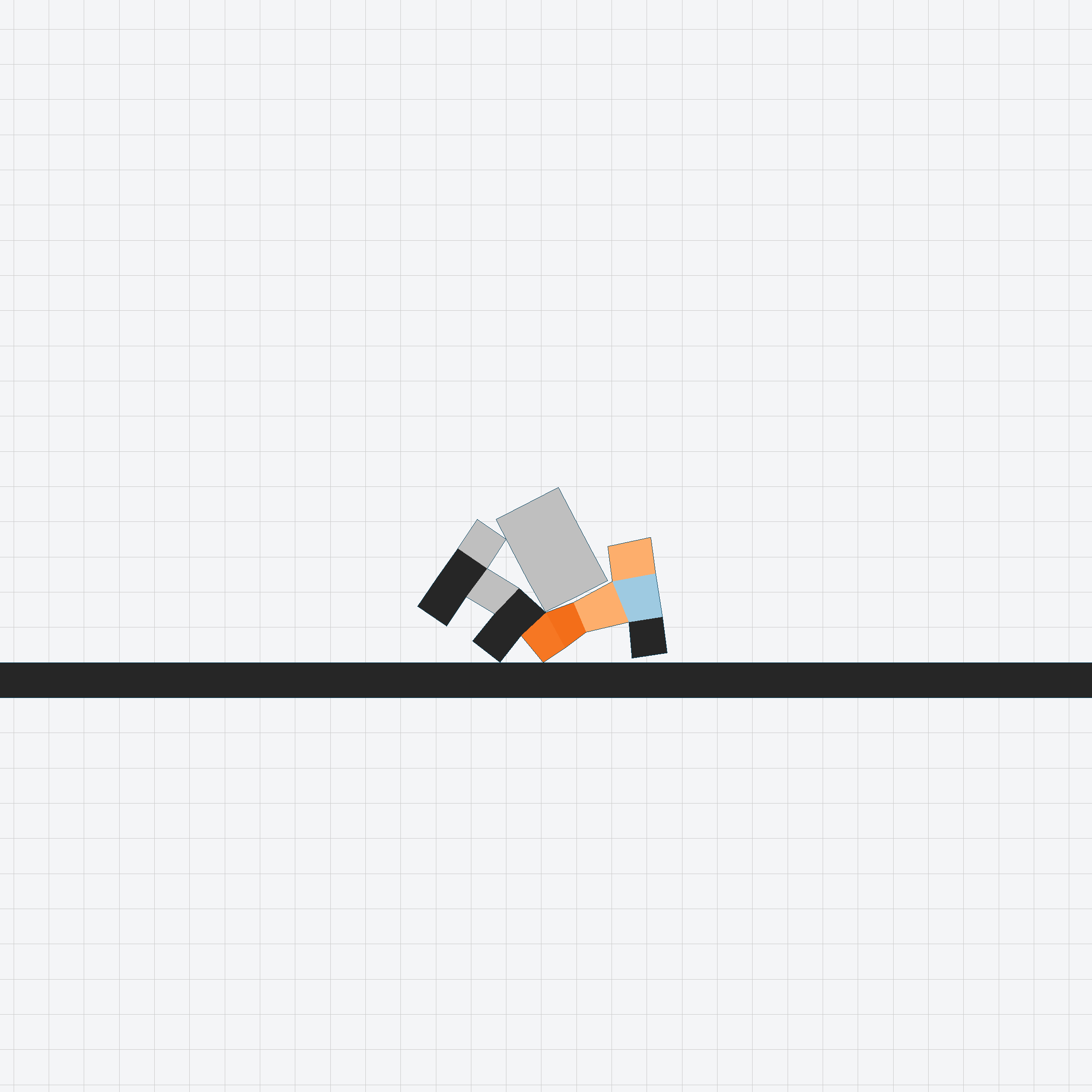}
    \adjincludegraphics[width=9mm,trim={{.25\width} {.25\width} {.25\width} {.25\width}},clip]{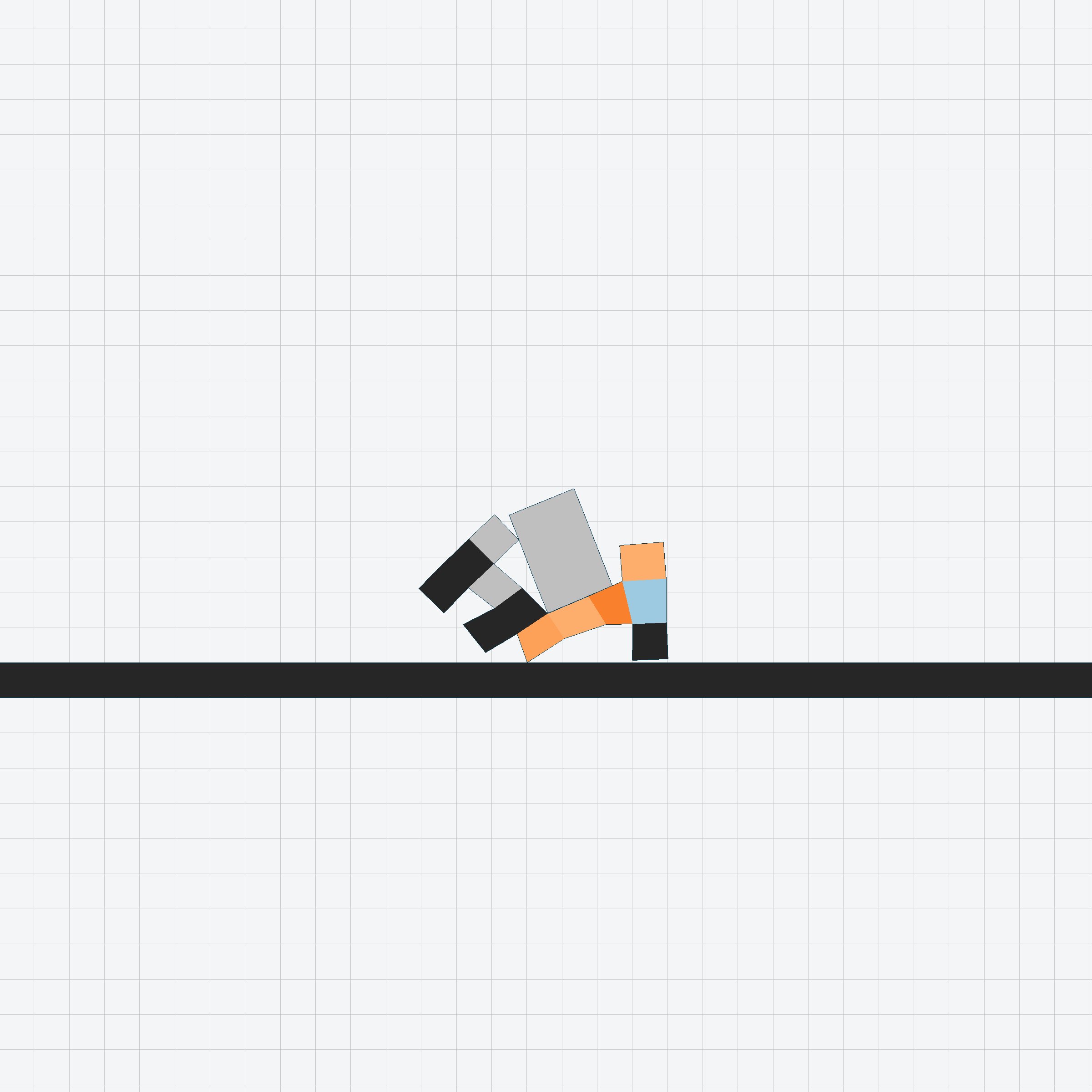}
    \adjincludegraphics[width=9mm,trim={{.25\width} {.25\width} {.25\width} {.25\width}},clip]{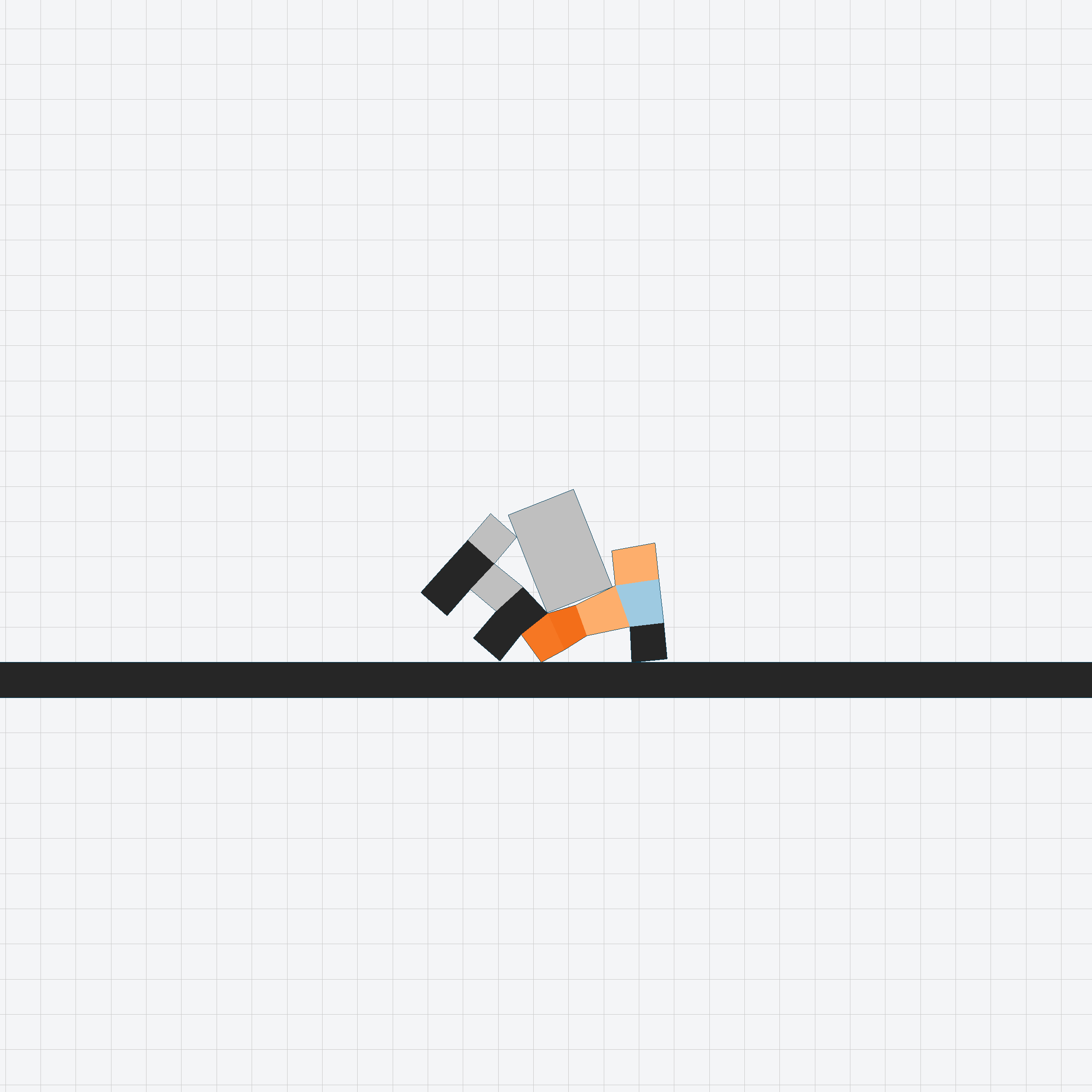}
    \adjincludegraphics[width=9mm,trim={{.25\width} {.25\width} {.25\width} {.25\width}},clip]{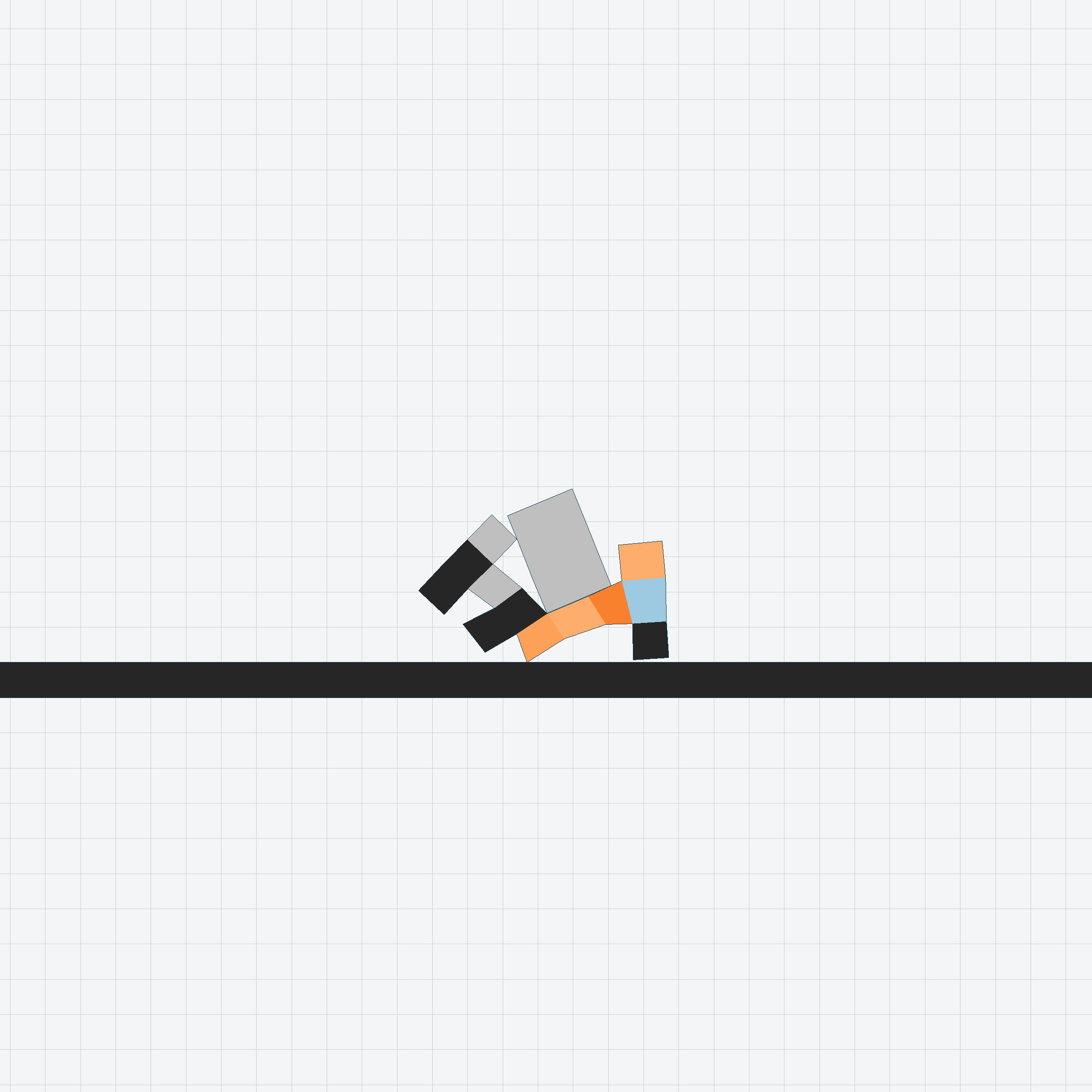}
    \adjincludegraphics[width=9mm,trim={{.25\width} {.25\width} {.25\width} {.25\width}},clip]{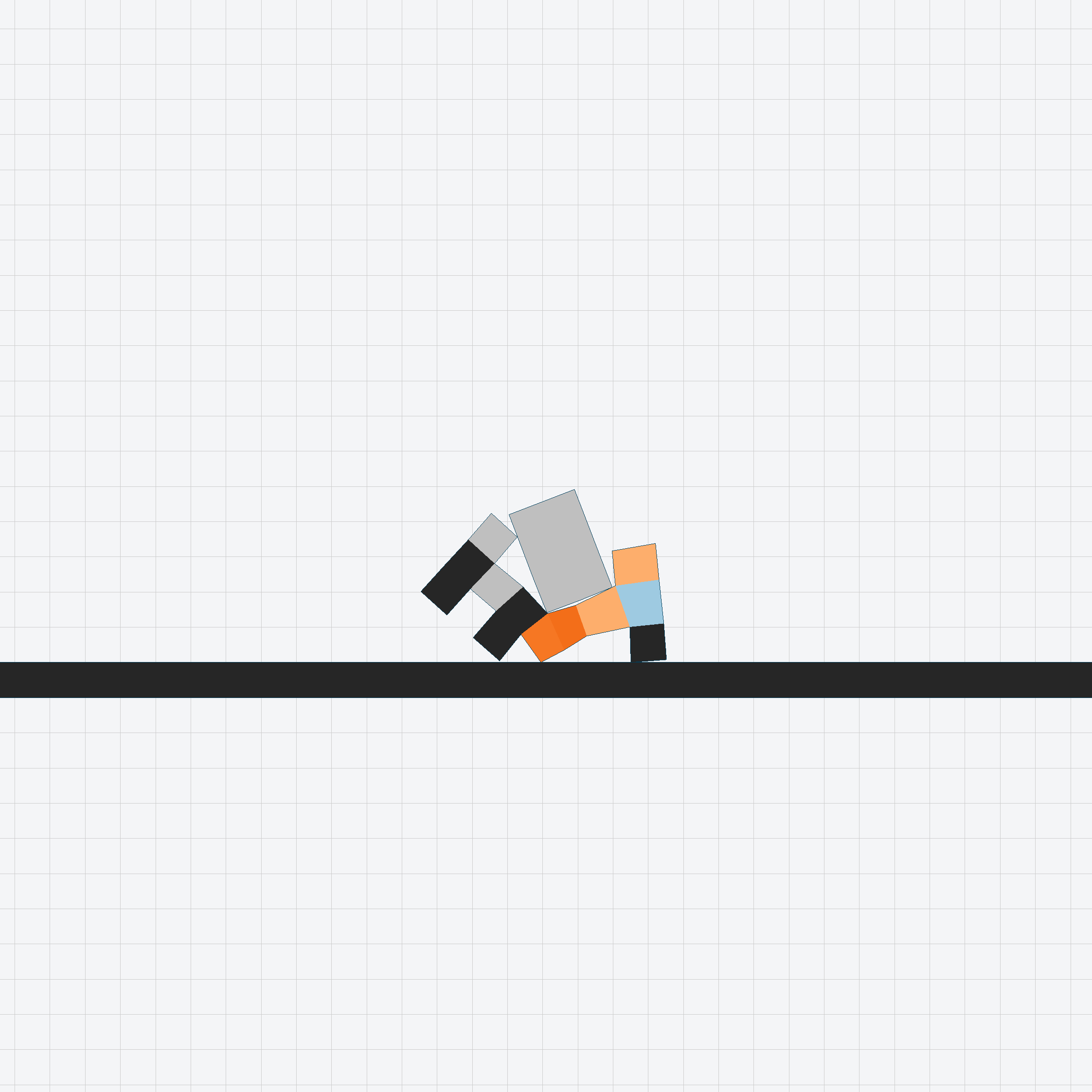}
    \caption{
        Eight snapshots of one episode of one of the \glspl{vsr} evolved with the Best-Many variant for the Catch task.
        The snapshots, taken at \qty{0.5}{\second} interval, show the robot grabbing and then carrying the payload.
    }
    \label{fig:catch-behavior-example}
\end{figure}

The figure shows that there are differences among \gls{vsr} bodies optimized for different tasks.
\glspl{vsr} evolved for locomotion are in general more compact, in particular for the Steps task.
Bodies optimized for carrying a payload (needed for both Carry and Catch) tend to have some protuberances (sort of ``arms'') for keeping the payload steady, which make the body less compact---see \Cref{fig:catch-behavior-example}.
This interpretation is qualitatively confirmed by \Cref{fig:best-morphologies}, where we show one random optimized \gls{vsr} body chosen among the $20$ repetitions for each approach-task combination.
On the other hand, \Cref{fig:body-features} highlights no differences among approaches: different kinds of learning do not lead to different bodies, at least according to these descriptors.

\begin{figure}
    \centering
    \begin{tabular}{l c@{\hspace{2mm}}c@{\hspace{2mm}}c@{\hspace{2mm}}c@{\hspace{2mm}}c@{\hspace{2mm}}c@{\hspace{2mm}}c@{\hspace{2mm}}c}
        \toprule
        & & & \multicolumn{2}{c}{Best} & \multicolumn{2}{c}{Similar} & \multicolumn{2}{c}{Random} \\
        \cmidrule(lr){4-5} \cmidrule(lr){6-7} \cmidrule(lr){8-9}
        Task & IL & Parent & One & Many & One & Many & One & Many \\
        \midrule
        Simple
        & \vsrevogym{5}{5}{00034-00040-00040-22140-03000}
        & \vsrevogym{5}{5}{40000-14000-10001-33331-00001}
        & \vsrevogym{5}{5}{20020-10424-33331-00001-00000}
        & \vsrevogym{5}{5}{00330-40312-10411-33334-00001}
        & \vsrevogym{5}{5}{42300-23410-10113-33334-00001}
        & \vsrevogym{5}{5}{20400-30210-30444-33334-00003}
        & \vsrevogym{5}{5}{00000-20024-30043-33334-00001}
        & \vsrevogym{5}{5}{00400-40110-40410-33334-00001}
        \\
        Steps
        & \vsrevogym{5}{5}{00002-12442-33334-00031-00001}
        & \vsrevogym{5}{5}{34444-23400-33400-01400-11430}
        & \vsrevogym{5}{5}{03010-01113-33334-00001-00000}
        & \vsrevogym{5}{5}{14341-44444-12340-41140-32240}
        & \vsrevogym{5}{5}{00031-00114-00024-03124-04204}
        & \vsrevogym{5}{5}{14113-32420-21400-43400-31400}
        & \vsrevogym{5}{5}{34123-30124-33333-00014-00002}
        & \vsrevogym{5}{5}{01332-33411-41400-11400-21400}
        \\
        Carry
        & \vsrevogym{5}{5}{21000-10000-20020-33321-00021}
        & \vsrevogym{5}{5}{00000-11001-20001-33331-00001}
        & \vsrevogym{5}{5}{00000-32003-20003-33331-00004}
        & \vsrevogym{5}{5}{20000-10023-40023-33333-00002}
        & \vsrevogym{5}{5}{00000-31000-40001-33332-00004}
        & \vsrevogym{5}{5}{04000-32000-20001-33333-00003}
        & \vsrevogym{5}{5}{13000-30003-30011-33331-00001}
        & \vsrevogym{5}{5}{04000-34004-40003-33333-00001}
        \\
        Catch
        & \vsrevogym{5}{5}{00000-22003-20001-33331-00002}
        & \vsrevogym{5}{5}{21000-30000-20023-33313-00001}
        & \vsrevogym{5}{5}{20000-10101-33333-00023-00000}
        & \vsrevogym{5}{5}{30000-12000-10000-33334-00004}
        & \vsrevogym{5}{5}{01000-14000-20001-33331-00010}
        & \vsrevogym{5}{5}{30000-30003-30003-33331-00004}
        & \vsrevogym{5}{5}{22000-14004-20011-33333-00022}
        & \vsrevogym{5}{5}{40000-20002-30032-33331-00014}
        \\
        \bottomrule
    \end{tabular}
    \caption{
        A selection of the optimized \gls{vsr} bodies, one per each approach (column) and task (row).
    }
    \label{fig:best-morphologies}
\end{figure}

\subsection{Potential and transferability of \gls{vsr} bodies}
\label{sec:exp-body-potential-transferability}
\begin{figure}
    \centering
    \begin{tikzpicture}
        \begin{groupplot}[
            boxplot,
            boxplot/draw direction=y,
            scale only axis,
            width=20mm,
            height=22.5mm,
            group style={
                group size=3 by 1,
                horizontal sep=5mm,
                vertical sep=5mm,
                xticklabels at=edge bottom,
                ylabels at=edge left,
            },
            ygridded,
            noinnerticks,
            ylabel={Quality~$q^\star$},
        ]
            \nextgroupplot[title={Simple}]
            \boxplotoutliers[bpcolor=col.individual]{plot-data/6/6-relearn-simple.txt}{Individual};
            \boxplotoutliers[bpcolor=col.parent]{plot-data/6/6-relearn-simple.txt}{Parent};
            \boxplotoutliers[bpcolor=col.best1]{plot-data/6/6-relearn-simple.txt}{Best-N=1};
            \boxplotoutliers[bpcolor=col.best8]{plot-data/6/6-relearn-simple.txt}{Best-N=8};
            \boxplotoutliers[bpcolor=col.similar1]{plot-data/6/6-relearn-simple.txt}{Similar-N=1};
            \boxplotoutliers[bpcolor=col.similar8]{plot-data/6/6-relearn-simple.txt}{Similar-N=8};
            \boxplotoutliers[bpcolor=col.random1]{plot-data/6/6-relearn-simple.txt}{Random-N=1};
            \boxplotoutliers[bpcolor=col.random8]{plot-data/6/6-relearn-simple.txt}{Random-N=8};
            
            \nextgroupplot[title={Steps}]
            \boxplotoutliers[bpcolor=col.individual]{plot-data/6/6-relearn-steps.txt}{Individual};
            \boxplotoutliers[bpcolor=col.parent]{plot-data/6/6-relearn-steps.txt}{Parent};
            \boxplotoutliers[bpcolor=col.best1]{plot-data/6/6-relearn-steps.txt}{Best-N=1};
            \boxplotoutliers[bpcolor=col.best8]{plot-data/6/6-relearn-steps.txt}{Best-N=8};
            \boxplotoutliers[bpcolor=col.similar1]{plot-data/6/6-relearn-steps.txt}{Similar-N=1};
            \boxplotoutliers[bpcolor=col.similar8]{plot-data/6/6-relearn-steps.txt}{Similar-N=8};
            \boxplotoutliers[bpcolor=col.random1]{plot-data/6/6-relearn-steps.txt}{Random-N=1};
            \boxplotoutliers[bpcolor=col.random8]{plot-data/6/6-relearn-steps.txt}{Random-N=8};

            \nextgroupplot[title={Carry}]
            \boxplotoutliers[bpcolor=col.individual]{plot-data/6/6-relearn-carry.txt}{Individual};
            \boxplotoutliers[bpcolor=col.parent]{plot-data/6/6-relearn-carry.txt}{Parent};
            \boxplotoutliers[bpcolor=col.best1]{plot-data/6/6-relearn-carry.txt}{Best-N=1};
            \boxplotoutliers[bpcolor=col.best8]{plot-data/6/6-relearn-carry.txt}{Best-N=8};
            \boxplotoutliers[bpcolor=col.similar1]{plot-data/6/6-relearn-carry.txt}{Similar-N=1};
            \boxplotoutliers[bpcolor=col.similar8]{plot-data/6/6-relearn-carry.txt}{Similar-N=8};
            \boxplotoutliers[bpcolor=col.random1]{plot-data/6/6-relearn-carry.txt}{Random-N=1};
            \boxplotoutliers[bpcolor=col.random8]{plot-data/6/6-relearn-carry.txt}{Random-N=8};
        \end{groupplot}
    \end{tikzpicture}
    \begin{tabular}{rrr}
        &
        Best-One \addlegendimageintext{boxplot legend image={col.best1}{}} &
        Best-Many \addlegendimageintext{boxplot legend image={col.best8}{}} \\
        \gls{il} \addlegendimageintext{boxplot legend image={col.individual}{}} &
        Similar-One \addlegendimageintext{boxplot legend image={col.similar1}{}} &
        Similar-Many \addlegendimageintext{boxplot legend image={col.similar8}{}} \\
        Parent \addlegendimageintext{boxplot legend image={col.parent}{}} &
        Random-One \addlegendimageintext{boxplot legend image={col.random1}{}} &
        Random-Many \addlegendimageintext{boxplot legend image={col.random8}{}}
    \end{tabular}
    \caption{
        Distribution of the quality $q^\star$ of the best-performing \gls{vsr} subjected to the re-learning of the brain with a large budget for \gls{bo}.
    }
    \label{fig:relearning-q-boxplots}
\end{figure}
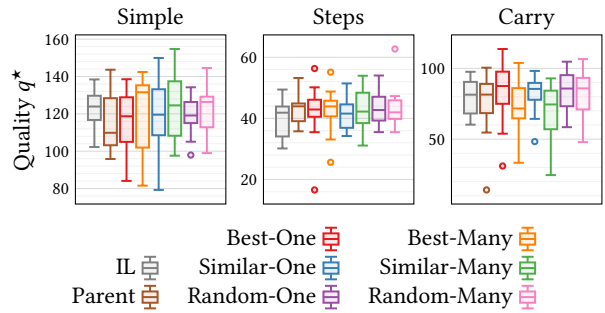

Having verified that different forms of learning do not impact on body diversity, nor on the general shape for the \gls{vsr} body (which is, instead, impacted by the task), we investigated the possibility that \glspl{vsr} obtained with different approaches have different \emph{potential for learning}.
Here, we use ``potential for learning'' to quantify the performance of a \gls{vsr} body with the ``best'' brain it can be paired with.
For this purpose, we took all the $20$ \glspl{vsr} obtained for each approach and task and, for each one, we \emph{re-learned} a new brain for the same task by using \gls{bo} with a random starting controller parameter set and reaching $n_\text{final}=500$ samples.
We assume that this learning budget for \gls{bo} is enough for revealing the full potential of any body.
We report in \Cref{fig:relearning-q-boxplots} the quality $q^\star$ of the \glspl{vsr} obtained with this procedure for the three static tasks.

It can be seen that there are no clear differences in the quality of the re-learned \glspl{vsr}: a performance difference which existed after the body-brain co-optimization can be nullified by pairing each body with its ``optimal'' brain.
A potential and practical consequence of this findings is that there are no reasons for not employing \gls{sl}, namely the Best-Many variant, when co-optimizing the body and the brain of a \gls{vsr}: robots obtained this way are in general better and their bodies are not less diverse, nor ``potentially'' worse.

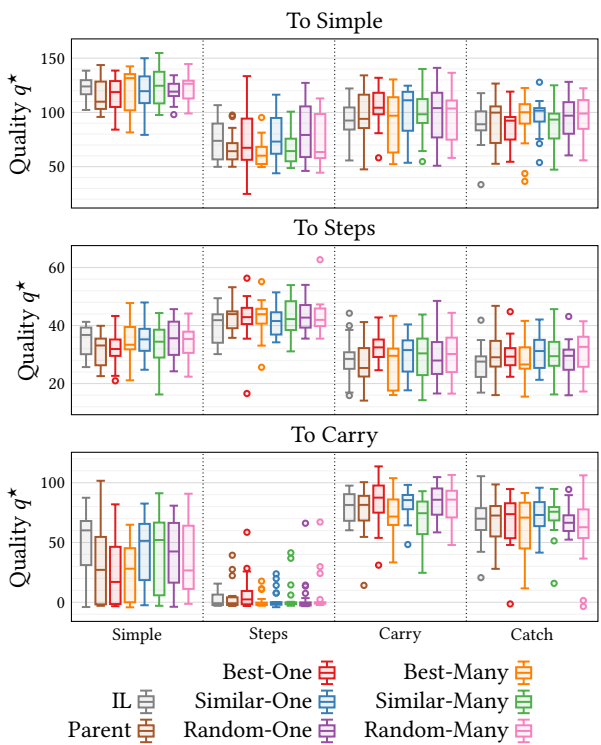
\begin{figure}
    \centering
    \begin{tikzpicture}
        \begin{groupplot}[
            scale only axis,
            groupedboxplot={8}{0.08},
            boxplot/draw direction=y,
            width=70mm,
            height=22.5mm,
            group style={
                group size=1 by 3,
                horizontal sep=1mm,
                vertical sep=5mm,
                xticklabels at=edge bottom,
                yticklabels at=edge left,
            },
            ygridded,
            noinnerticks,
            ylabel={Quality $q^\star$},
            xtick={1,2,3,4},
            xticklabels={Simple,Steps,Carry,Catch},
            xmin=.5,
            xmax=4.5
        ]
            \nextgroupplot[title={To Simple}]
                
                \boxplotoutliers[bpcolor=col.individual]{plot-data/11/11-relearn-simple-to-simple.txt}{Individual};
                \boxplotoutliers[bpcolor=col.parent]{plot-data/11/11-relearn-simple-to-simple.txt}{Parent};
                \boxplotoutliers[bpcolor=col.best1]{plot-data/11/11-relearn-simple-to-simple.txt}{Best-N=1};
                \boxplotoutliers[bpcolor=col.best8]{plot-data/11/11-relearn-simple-to-simple.txt}{Best-N=8};
                \boxplotoutliers[bpcolor=col.similar1]{plot-data/11/11-relearn-simple-to-simple.txt}{Similar-N=1};
                \boxplotoutliers[bpcolor=col.similar8]{plot-data/11/11-relearn-simple-to-simple.txt}{Similar-N=8};
                \boxplotoutliers[bpcolor=col.random1]{plot-data/11/11-relearn-simple-to-simple.txt}{Random-N=1};
                \boxplotoutliers[bpcolor=col.random8]{plot-data/11/11-relearn-simple-to-simple.txt}{Random-N=8};

                \boxplotoutliers[bpcolor=col.individual]{plot-data/11/11-relearn-steps-to-simple.txt}{Individual};
                \boxplotoutliers[bpcolor=col.parent]{plot-data/11/11-relearn-steps-to-simple.txt}{Parent};
                \boxplotoutliers[bpcolor=col.best1]{plot-data/11/11-relearn-steps-to-simple.txt}{Best-N=1};
                \boxplotoutliers[bpcolor=col.best8]{plot-data/11/11-relearn-steps-to-simple.txt}{Best-N=8};
                \boxplotoutliers[bpcolor=col.similar1]{plot-data/11/11-relearn-steps-to-simple.txt}{Similar-N=1};
                \boxplotoutliers[bpcolor=col.similar8]{plot-data/11/11-relearn-steps-to-simple.txt}{Similar-N=8};
                \boxplotoutliers[bpcolor=col.random1]{plot-data/11/11-relearn-steps-to-simple.txt}{Random-N=1};
                \boxplotoutliers[bpcolor=col.random8]{plot-data/11/11-relearn-steps-to-simple.txt}{Random-N=8};

                \boxplotoutliers[bpcolor=col.individual]{plot-data/11/11-relearn-carry-to-simple.txt}{Individual};
                \boxplotoutliers[bpcolor=col.parent]{plot-data/11/11-relearn-carry-to-simple.txt}{Parent};
                \boxplotoutliers[bpcolor=col.best1]{plot-data/11/11-relearn-carry-to-simple.txt}{Best-N=1};
                \boxplotoutliers[bpcolor=col.best8]{plot-data/11/11-relearn-carry-to-simple.txt}{Best-N=8};
                \boxplotoutliers[bpcolor=col.similar1]{plot-data/11/11-relearn-carry-to-simple.txt}{Similar-N=1};
                \boxplotoutliers[bpcolor=col.similar8]{plot-data/11/11-relearn-carry-to-simple.txt}{Similar-N=8};
                \boxplotoutliers[bpcolor=col.random1]{plot-data/11/11-relearn-carry-to-simple.txt}{Random-N=1};
                \boxplotoutliers[bpcolor=col.random8]{plot-data/11/11-relearn-carry-to-simple.txt}{Random-N=8};

                \boxplotoutliers[bpcolor=col.individual]{plot-data/11/11-relearn-catch-to-simple.txt}{Individual};
                \boxplotoutliers[bpcolor=col.parent]{plot-data/11/11-relearn-catch-to-simple.txt}{Parent};
                \boxplotoutliers[bpcolor=col.best1]{plot-data/11/11-relearn-catch-to-simple.txt}{Best-N=1};
                \boxplotoutliers[bpcolor=col.best8]{plot-data/11/11-relearn-catch-to-simple.txt}{Best-N=8};
                \boxplotoutliers[bpcolor=col.similar1]{plot-data/11/11-relearn-catch-to-simple.txt}{Similar-N=1};
                \boxplotoutliers[bpcolor=col.similar8]{plot-data/11/11-relearn-catch-to-simple.txt}{Similar-N=8};
                \boxplotoutliers[bpcolor=col.random1]{plot-data/11/11-relearn-catch-to-simple.txt}{Random-N=1};
                \boxplotoutliers[bpcolor=col.random8]{plot-data/11/11-relearn-catch-to-simple.txt}{Random-N=8};

                \draw [densely dotted] (1.5,\pgfkeysvalueof{/pgfplots/ymin}) -- (1.5,\pgfkeysvalueof{/pgfplots/ymax});
                \draw [densely dotted] (2.5,\pgfkeysvalueof{/pgfplots/ymin}) -- (2.5,\pgfkeysvalueof{/pgfplots/ymax});
                \draw [densely dotted] (3.5,\pgfkeysvalueof{/pgfplots/ymin}) -- (3.5,\pgfkeysvalueof{/pgfplots/ymax});

            \nextgroupplot[title={To Steps}]
                
                \boxplotoutliers[bpcolor=col.individual]{plot-data/11/11-relearn-simple-to-steps.txt}{Individual};
                \boxplotoutliers[bpcolor=col.parent]{plot-data/11/11-relearn-simple-to-steps.txt}{Parent};
                \boxplotoutliers[bpcolor=col.best1]{plot-data/11/11-relearn-simple-to-steps.txt}{Best-N=1};
                \boxplotoutliers[bpcolor=col.best8]{plot-data/11/11-relearn-simple-to-steps.txt}{Best-N=8};
                \boxplotoutliers[bpcolor=col.similar1]{plot-data/11/11-relearn-simple-to-steps.txt}{Similar-N=1};
                \boxplotoutliers[bpcolor=col.similar8]{plot-data/11/11-relearn-simple-to-steps.txt}{Similar-N=8};
                \boxplotoutliers[bpcolor=col.random1]{plot-data/11/11-relearn-simple-to-steps.txt}{Random-N=1};
                \boxplotoutliers[bpcolor=col.random8]{plot-data/11/11-relearn-simple-to-steps.txt}{Random-N=8};

                \boxplotoutliers[bpcolor=col.individual]{plot-data/11/11-relearn-steps-to-steps.txt}{Individual};
                \boxplotoutliers[bpcolor=col.parent]{plot-data/11/11-relearn-steps-to-steps.txt}{Parent};
                \boxplotoutliers[bpcolor=col.best1]{plot-data/11/11-relearn-steps-to-steps.txt}{Best-N=1};
                \boxplotoutliers[bpcolor=col.best8]{plot-data/11/11-relearn-steps-to-steps.txt}{Best-N=8};
                \boxplotoutliers[bpcolor=col.similar1]{plot-data/11/11-relearn-steps-to-steps.txt}{Similar-N=1};
                \boxplotoutliers[bpcolor=col.similar8]{plot-data/11/11-relearn-steps-to-steps.txt}{Similar-N=8};
                \boxplotoutliers[bpcolor=col.random1]{plot-data/11/11-relearn-steps-to-steps.txt}{Random-N=1};
                \boxplotoutliers[bpcolor=col.random8]{plot-data/11/11-relearn-steps-to-steps.txt}{Random-N=8};

                \boxplotoutliers[bpcolor=col.individual]{plot-data/11/11-relearn-carry-to-steps.txt}{Individual};
                \boxplotoutliers[bpcolor=col.parent]{plot-data/11/11-relearn-carry-to-steps.txt}{Parent};
                \boxplotoutliers[bpcolor=col.best1]{plot-data/11/11-relearn-carry-to-steps.txt}{Best-N=1};
                \boxplotoutliers[bpcolor=col.best8]{plot-data/11/11-relearn-carry-to-steps.txt}{Best-N=8};
                \boxplotoutliers[bpcolor=col.similar1]{plot-data/11/11-relearn-carry-to-steps.txt}{Similar-N=1};
                \boxplotoutliers[bpcolor=col.similar8]{plot-data/11/11-relearn-carry-to-steps.txt}{Similar-N=8};
                \boxplotoutliers[bpcolor=col.random1]{plot-data/11/11-relearn-carry-to-steps.txt}{Random-N=1};
                \boxplotoutliers[bpcolor=col.random8]{plot-data/11/11-relearn-carry-to-steps.txt}{Random-N=8};

                \boxplotoutliers[bpcolor=col.individual]{plot-data/11/11-relearn-catch-to-steps.txt}{Individual};
                \boxplotoutliers[bpcolor=col.parent]{plot-data/11/11-relearn-catch-to-steps.txt}{Parent};
                \boxplotoutliers[bpcolor=col.best1]{plot-data/11/11-relearn-catch-to-steps.txt}{Best-N=1};
                \boxplotoutliers[bpcolor=col.best8]{plot-data/11/11-relearn-catch-to-steps.txt}{Best-N=8};
                \boxplotoutliers[bpcolor=col.similar1]{plot-data/11/11-relearn-catch-to-steps.txt}{Similar-N=1};
                \boxplotoutliers[bpcolor=col.similar8]{plot-data/11/11-relearn-catch-to-steps.txt}{Similar-N=8};
                \boxplotoutliers[bpcolor=col.random1]{plot-data/11/11-relearn-catch-to-steps.txt}{Random-N=1};
                \boxplotoutliers[bpcolor=col.random8]{plot-data/11/11-relearn-catch-to-steps.txt}{Random-N=8};

                \draw [densely dotted] (1.5,\pgfkeysvalueof{/pgfplots/ymin}) -- (1.5,\pgfkeysvalueof{/pgfplots/ymax});
                \draw [densely dotted] (2.5,\pgfkeysvalueof{/pgfplots/ymin}) -- (2.5,\pgfkeysvalueof{/pgfplots/ymax});
                \draw [densely dotted] (3.5,\pgfkeysvalueof{/pgfplots/ymin}) -- (3.5,\pgfkeysvalueof{/pgfplots/ymax});

            \nextgroupplot[title={To Carry}]
                
                \boxplotoutliers[bpcolor=col.individual]{plot-data/11/11-relearn-simple-to-carry.txt}{Individual};
                \boxplotoutliers[bpcolor=col.parent]{plot-data/11/11-relearn-simple-to-carry.txt}{Parent};
                \boxplotoutliers[bpcolor=col.best1]{plot-data/11/11-relearn-simple-to-carry.txt}{Best-N=1};
                \boxplotoutliers[bpcolor=col.best8]{plot-data/11/11-relearn-simple-to-carry.txt}{Best-N=8};
                \boxplotoutliers[bpcolor=col.similar1]{plot-data/11/11-relearn-simple-to-carry.txt}{Similar-N=1};
                \boxplotoutliers[bpcolor=col.similar8]{plot-data/11/11-relearn-simple-to-carry.txt}{Similar-N=8};
                \boxplotoutliers[bpcolor=col.random1]{plot-data/11/11-relearn-simple-to-carry.txt}{Random-N=1};
                \boxplotoutliers[bpcolor=col.random8]{plot-data/11/11-relearn-simple-to-carry.txt}{Random-N=8};

                \boxplotoutliers[bpcolor=col.individual]{plot-data/11/11-relearn-steps-to-carry.txt}{Individual};
                \boxplotoutliers[bpcolor=col.parent]{plot-data/11/11-relearn-steps-to-carry.txt}{Parent};
                \boxplotoutliers[bpcolor=col.best1]{plot-data/11/11-relearn-steps-to-carry.txt}{Best-N=1};
                \boxplotoutliers[bpcolor=col.best8]{plot-data/11/11-relearn-steps-to-carry.txt}{Best-N=8};
                \boxplotoutliers[bpcolor=col.similar1]{plot-data/11/11-relearn-steps-to-carry.txt}{Similar-N=1};
                \boxplotoutliers[bpcolor=col.similar8]{plot-data/11/11-relearn-steps-to-carry.txt}{Similar-N=8};
                \boxplotoutliers[bpcolor=col.random1]{plot-data/11/11-relearn-steps-to-carry.txt}{Random-N=1};
                \boxplotoutliers[bpcolor=col.random8]{plot-data/11/11-relearn-steps-to-carry.txt}{Random-N=8};

                \boxplotoutliers[bpcolor=col.individual]{plot-data/11/11-relearn-carry-to-carry.txt}{Individual};
                \boxplotoutliers[bpcolor=col.parent]{plot-data/11/11-relearn-carry-to-carry.txt}{Parent};
                \boxplotoutliers[bpcolor=col.best1]{plot-data/11/11-relearn-carry-to-carry.txt}{Best-N=1};
                \boxplotoutliers[bpcolor=col.best8]{plot-data/11/11-relearn-carry-to-carry.txt}{Best-N=8};
                \boxplotoutliers[bpcolor=col.similar1]{plot-data/11/11-relearn-carry-to-carry.txt}{Similar-N=1};
                \boxplotoutliers[bpcolor=col.similar8]{plot-data/11/11-relearn-carry-to-carry.txt}{Similar-N=8};
                \boxplotoutliers[bpcolor=col.random1]{plot-data/11/11-relearn-carry-to-carry.txt}{Random-N=1};
                \boxplotoutliers[bpcolor=col.random8]{plot-data/11/11-relearn-carry-to-carry.txt}{Random-N=8};

                \boxplotoutliers[bpcolor=col.individual]{plot-data/11/11-relearn-catch-to-carry.txt}{Individual};
                \boxplotoutliers[bpcolor=col.parent]{plot-data/11/11-relearn-catch-to-carry.txt}{Parent};
                \boxplotoutliers[bpcolor=col.best1]{plot-data/11/11-relearn-catch-to-carry.txt}{Best-N=1};
                \boxplotoutliers[bpcolor=col.best8]{plot-data/11/11-relearn-catch-to-carry.txt}{Best-N=8};
                \boxplotoutliers[bpcolor=col.similar1]{plot-data/11/11-relearn-catch-to-carry.txt}{Similar-N=1};
                \boxplotoutliers[bpcolor=col.similar8]{plot-data/11/11-relearn-catch-to-carry.txt}{Similar-N=8};
                \boxplotoutliers[bpcolor=col.random1]{plot-data/11/11-relearn-catch-to-carry.txt}{Random-N=1};
                \boxplotoutliers[bpcolor=col.random8]{plot-data/11/11-relearn-catch-to-carry.txt}{Random-N=8};

                \draw [densely dotted] (1.5,\pgfkeysvalueof{/pgfplots/ymin}) -- (1.5,\pgfkeysvalueof{/pgfplots/ymax});
                \draw [densely dotted] (2.5,\pgfkeysvalueof{/pgfplots/ymin}) -- (2.5,\pgfkeysvalueof{/pgfplots/ymax});
                \draw [densely dotted] (3.5,\pgfkeysvalueof{/pgfplots/ymin}) -- (3.5,\pgfkeysvalueof{/pgfplots/ymax});

        \end{groupplot}
    \end{tikzpicture}
    \begin{tabular}{rrr}
        &
        Best-One \addlegendimageintext{boxplot legend image={col.best1}{}} &
        Best-Many \addlegendimageintext{boxplot legend image={col.best8}{}} \\
        \gls{il} \addlegendimageintext{boxplot legend image={col.individual}{}} &
        Similar-One \addlegendimageintext{boxplot legend image={col.similar1}{}} &
        Similar-Many \addlegendimageintext{boxplot legend image={col.similar8}{}} \\
        Parent \addlegendimageintext{boxplot legend image={col.parent}{}} &
        Random-One \addlegendimageintext{boxplot legend image={col.random1}{}} &
        Random-Many \addlegendimageintext{boxplot legend image={col.random8}{}}
    \end{tabular}
    \caption{
        Distribution of the quality $q^\star$ of the best-performing \gls{vsr} optimized on a source task (row of plots) subjected to the re-learning of the brain with a large budget for \gls{bo} and for another destination task (groups of boxes).
    }    
    \label{fig:transfer}
    \vspace{-5mm}
\end{figure}

We also measured the performance of \glspl{vsr} subjected to brain re-learning, but using another task for re-learning than the one used for evolving the body.
That is, we assessed the transferability of bodies to other tasks.
We limited this analysis to the three static tasks as destination task.
We performed the re-learning as in the previous experiment.
We show the results in \Cref{fig:transfer}.
For easing the comparison of performance of transferred bodies, we also report the re-learning results for the destination task being the same of the source one, \ie, the one where bodies have been evolved.

The main finding is that there is always a degradation of the performance after transferring the \glspl{vsr} to another task: this is more than sound, as different tasks ask for different bodies, but here we are just re-learning the brains.
The second finding is that, consistent with the previous experiment, re-learning makes the differences among \gls{sl} approaches vanish.

\subsection{Effectiveness of \gls{bo} as learning}
To validate the effectiveness of \gls{bo} and \gls{sl} as learning algorithms, we compared them to the No-\gls{bo} approach on a selection of fixed bodies.
We do this outside of the main evolutionary loop to isolate the effect of learning, \ie, without optimizing the body.
We used the eight selected optimized bodies for the Simple environment shown in the top row of \Cref{fig:best-morphologies}, and similarly as described in \Cref{sec:exp-body-potential-transferability}, we \emph{re-learned} a new brain from scratch using three approaches.
The three approaches are:
\begin{enumerate*}[(a)]
    \item a No-\gls{bo} baseline where all candidate controllers are sampled randomly;
    \item an \gls{il} approach; and
    \item a \gls{sl} approach where the transferred samples are taken from the \gls{il} approach (on the same body).
\end{enumerate*}
Note that for the \gls{sl} approach, the samples are only transferred from the first $50$ samples from the \gls{il} approach. 

\begin{figure}
    \centering
    \begin{tikzpicture}
        \begin{axis}[
            scale only axis,
            width=70mm,
            height=25mm,
            gridded,
            noinnerticks,
            xlabel={Learning iteration},
            ylabel={Quality $q^\star$},
            ymin=-10,
            ymax=140
        ]
            \addplot[densely dotted, red] coordinates {(50,-20) (50,140)};
            \lineminmax[lcolor=black]{plot-data/bo-effectiveness/individual.txt}{}{x}{y}{ymin}{ymax}            
            \lineminmax[lcolor=gray]{plot-data/bo-effectiveness/random.txt}{}{x}{y}{ymin}{ymax}
            \lineminmax[lcolor=col21]{plot-data/bo-effectiveness/parent-same.txt}{}{x}{y}{ymin}{ymax}
        \end{axis}
    \end{tikzpicture}
    No-\gls{bo} \addlegendimageintext{shaded legend image={gray}{solid}} \hspace{3mm}
    \gls{il} \addlegendimageintext{shaded legend image={black}{solid}} \hspace{3mm}
    \gls{sl} \addlegendimageintext{shaded legend image={col21}{solid}}
    \caption{
        Performance over learning iterations for three brain learning approaches evaluated on eight fixed robot bodies.
        The vertical red dotted line corresponds to the learning budget of $50$ in our main experiment.
    }
    \label{fig:bo-learning-effectiveness}
\end{figure}
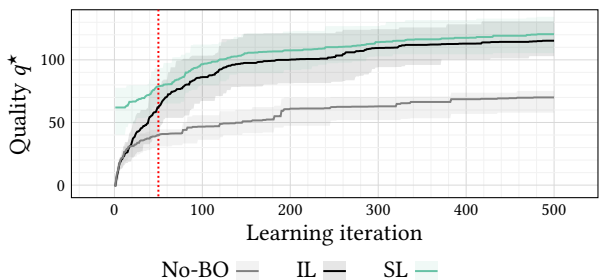

The results for this experiment are presented in \Cref{fig:bo-learning-effectiveness}.
Firstly, the results show that both \gls{bo} approaches quickly perform better than the No-\gls{bo} baseline, showing the effectiveness of \gls{bo}.
Comparing \gls{sl} to \gls{il}, we see that \gls{sl} has an expected initial advantage in performance, but over many learning iterations \gls{il} is able to close the gap in performance. 
This shows a clear benefit of \gls{sl} over \gls{il}, namely that it requires fewer learning iterations for good performance, making room for evaluating more robot bodies with the same number of total function evaluations, \ie, simulated episodes.

While our \gls{bo} approach is effective in the considered setting, it could be improved by using techniques from high-dimensional \gls{bo}.
For example, random embedding methods~\cite{wang2016bayesian}, trust-region-based BO methods such as TuRBO~\cite{eriksson2019scalable} and additive Gaussian process models~\cite{kandasamy2015high}.
We leave this to future work to investigate.

\section{Conclusion}
We have investigated \gls{sl} strategies to transfer information across generations in evolvable robots on four different environments and tasks.
In our work, the brain is optimized using \gls{bo} for each robot.
\Gls{sl} is done by transferring samples between robots, and we compare four different strategies: transferring samples from the best, the most similar, a random robot from the previous generation, and the parent.
We also look at the effects of having a single or multiple teachers.
We compare \gls{sl} strategies against \gls{il} where no samples are transferred between robots.

We consider four tasks where \glspl{vsr} need to be optimized in body and brain to perform movements and to carry payloads.
For all environments, the results show that often any \gls{sl} strategy outperforms \gls{il}.
Among the \gls{sl} strategies, learning from the best robot often performs better than other strategies.
Future work can look into the effect diversity-promoting algorithms on the \gls{sl} strategies.
Moreover, learning from more teachers is generally better than from one teacher, despite the amount of information being the same.

We do not see a noticeable difference in morphologies and morphological diversity between strategies.
Also, we show that a difference in performance after co-evolving the \emph{body and the brain} does not necessarily mean that the \emph{body} has a better ability to learn the task.
Therefore, the reason that, during co-evolution, the individuals for a strategy perform better might be only because there are better control parameters found.
This is true for both re-learning the brains of the optimized bodies in the \emph{same} environment and task, or when bodies are transferred to \emph{different} environments.
If the goal is to end up with a single good-performing robot body, there is less benefit of \gls{sl}.
However, in a continuous evolutionary cycle, our \gls{sl} approaches are good strategies, since better control parameters are found for the individuals.
Related to this, a dynamic environment still showed benefits of \gls{sl}, despite related work showing diminishing benefits.
Future work should look deeper into this, for example, with varying degrees of changes in environments.

\section{Acknowledgements}

The paper is based upon work from a scholarship supported by SPECIES (http://species-society.org), the Society for the Promotion of Evolutionary Computation in Europe and its Surroundings. The work was also supported by The Research Council of Norway (RCN) through its Center of Excellence scheme, RITMO with Project No. 262762, and through the Norwegian Center for Embodied AI (NCEI) under grant agreement no. 357451

\bibliographystyle{ACM-Reference-Format}

\end{document}